\pdfoutput=1

\documentclass[11pt]{article}

\usepackage{ACL2023}

\usepackage{times}
\usepackage{latexsym}

\usepackage[T1]{fontenc}

\usepackage[utf8]{inputenc}

\usepackage{microtype}

\usepackage{inconsolata}

\usepackage{xspace}
\usepackage[frozencache,cachedir=.]{minted}
\usepackage{booktabs}
\usepackage{amsmath}
\usepackage{amssymb}
\usepackage{graphicx}
\usepackage{adjustbox}
\usepackage{multirow}
\usepackage{comment}
\usepackage{color, colortbl}
\usepackage{arydshln}
\definecolor{Gray}{gray}{0.9}
\newcolumntype{g}{>{\columncolor{Gray}}c}
\newcolumntype{h}{>{\columncolor{Gray}}l}

\newcommand{\rparagraph}[1]{\vspace{1.2mm}\noindent\textbf{#1.}}
\newcommand{\sparagraph}[1]{\vspace{0.0mm}\noindent\textbf{#1.}}

\newcommand{\zsxlt}{{\textsc{zs-xlt}}\xspace}
\newcommand{\fsxlt}{{\textsc{fs-xlt}}\xspace}
\newcommand{\xlt}{{\textsc{xlt}}\xspace}

\newcommand{\last}{{\textsc{last}}\xspace}
\newcommand{\srcdev}{{\textsc{src-dev}}\xspace}
\newcommand{\trgdev}{{\textsc{trg-dev}}\xspace}
\newcommand{\sdev}{{\textsc{s-dev}}\xspace}
\newcommand{\tdev}{{\textsc{t-dev}}\xspace}

\newcommand{\ca}{{\textsc{ca}}\xspace}
\newcommand{\ra}{{\textsc{ra}}\xspace}
\newcommand{\raca}{{\textsc{ra-ca}}\xspace}
\newcommand{\ralast}{{\textsc{ra-last}}\xspace}
\newcommand{\soupca}{{\textsc{soup-ca}}\xspace}
\newcommand{\souplast}{{\textsc{soup-last}}\xspace}

\newcommand{\gs}{{\textsc{gs}}\xspace}

\title{\textit{Free Lunch}: Robust Cross-Lingual Transfer \\ via Model Checkpoint Averaging}

\author{Fabian David Schmidt\textsuperscript{1}, Ivan Vulić\textsuperscript{2}, Goran Glavaš\textsuperscript{1} \\
  \textsuperscript{1} Center For Artificial Intelligence and Data Science, University of Würzburg, Germany \\
  \textsuperscript{2} Language Technology Lab, University of Cambridge, UK \\
  \texttt{\{fabian.schmidt, goran.glavas\}@uni-wuerzburg.de} \\
  \texttt{iv250@cam.ac.uk} }

\begin{document}
\maketitle
\begin{abstract}
Massively multilingual language models have displayed strong performance in zero-shot (\zsxlt) and few-shot (\fsxlt) cross-lingual transfer setups, where models fine-tuned on task data in a source language are transferred without any or with only a few annotated instances to the target language(s).
However, current work typically overestimates model performance as fine-tuned models are frequently evaluated at model checkpoints that generalize best to validation instances in the target languages.
This effectively violates the main assumptions of \textit{`true'} \zsxlt and \fsxlt.
Such \xlt setups require robust methods that do not depend on labeled target language data for validation and model selection.
In this work, aiming to improve the robustness of `true' \zsxlt and \fsxlt, we propose a simple and effective method that \textit{averages different checkpoints} (i.e., model snapshots) during task fine-tuning. 
We conduct exhaustive \zsxlt and \fsxlt experiments across higher-level semantic tasks (NLI, extractive QA) and lower-level token classification tasks (NER, POS).
The results indicate that averaging model checkpoints yields systematic and consistent performance gains across diverse target languages in all tasks.
Importantly, it simultaneously substantially desensitizes \xlt to varying hyperparameter choices in the absence of target language validation.
We also show that checkpoint averaging benefits performance 
when further combined with {\it run averaging} (i.e., averaging the parameters of models fine-tuned over independent runs).
\end{abstract}

\section{Introduction and Motivation}
\label{sec:intro}
Massively multilingual transformers (MMT) such as mBERT \citep{devlin-etal-2019-bert} and XLM-R \citep{conneau-etal-xlmr} have become the main driver of multilingual NLP research.
When fine-tuned on sizable task data in a high-resource source language, typically English, MMTs demonstrate cross-lingual transfer capabilities \citep{pires-etal-2019-multilingual} in \textit{zero-shot} (\zsxlt; without any task-annotated instances in the target language) and \textit{few-shot} (\fsxlt; only a few task-annotated instances/shots available in the target language) transfer setups~\citep{hu2020xtreme, lauscher-etal-2020-zero}.
However, recent work has shown that both cross-lingual transfer (\xlt) paradigms are subject to large variation in \xlt performance, especially if the target language is typologically distant to the source~\citep{keung-etal-2020-dont,zhao-etal-2021-closer, schmidt2022dontstopft}.

The protocols for model selection in previous \xlt work vary broadly, which exacerbates the comparison of reported \xlt results. Some studies (i) do not sufficiently discuss their protocol~\citep{conneau-etal-xlmr,xu-etal-2022-s4}, while others (ii) tune hyperparameters on the English development splits~\citep{hu2020xtreme,wu-dredze-2020-explicit}, or even (iii) perform model selection on the target-language validation sets~\citep{luo-etal-2021-veco, Fang_Wang_Gan_Sun_Liu_2021, zhao-etal-2021-closer}.
Assuming the availability of sufficiently large target-language validation sets for hyperparameter-tuning and model selection is unrealistic and violates the assumption of a true \zsxlt and \fsxlt setup~\cite{Perez:2021neurips,schmidt2022dontstopft}. On the other hand, model selection on English validation data often does not correlate well with target-language performance~\citep{keung-etal-2020-dont}.

Furthermore, benchmarking new and emerging \xlt approaches with existing methods is even more challenging when the code or models from prior work are not publicly available (e.g., \citealp{wei2021on, xu-etal-2022-s4}).\footnote{Even when they are available, conducting comparative evaluations incurs an overhead of navigating an unfamiliar code base and potentially higher runtime.} We therefore seek methods that reliably improve \zsxlt and \fsxlt irrespective of the underlying model and the transfer paradigm, are easy to implement, inexpensive to evaluate, robust to varying hyperparameters, and applicable to \textit{true} \xlt setups where the existence of any target-language validation data cannot be assumed nor guaranteed. 

In this work, we propose a simple and effective method of \textit{checkpoint averaging} (\ca) that satisfies all the desiderata above. 
The principal idea is to save \textit{model snapshots} at periodic intervals during fine-tuning and then average the weights of the multiple single-run snapshots (i.e., checkpoints) prior to \xlt evaluation. 
A similar procedure has been successfully adopted, for instance, in computer vision \citep{huang2017snapshot}, other NLP domains such as machine translation (\citealp{vaswani2017nips, gao-etal-2022-revisiting}, \textit{inter alia}), and speech processing (\citealp{dong_speech,Karita2019ACS}, \textit{inter alia}); however, it has not investigated nor adequately leveraged in \xlt, notorious for its sensitivity to different choices of shots and hyperparameters.

Averaging model weights can be extended to merging last or multiple model snapshots from \textit{multiple model runs} in a straightforward manner. As we show later, within-run snapshot averaging performs comparable, or even better in individual experiments, than the computationally more expensive ensembling of last snapshots of multiple models (i.e., from different training runs). 

\rparagraph{Contributions}
\textbf{(1)} To the best of our knowledge, we are the first to extensively benchmark and analyze \ca for both \zsxlt and \fsxlt; we do this on a range of higher-level semantic (NLI, extractive QA) and lower-level token classification tasks (NER, POS).
\ca yields two benefits in true \xlt setups, coming for `free' (i.e., at no additional computation cost):
the transfer performance (i) improves consistently, and (ii) it becomes much less sensitive to varying hyperparameters.
\textbf{(2)} We shed more light on averaging models across runs (i.e., ensembling). We first confirm that standard plain ensembling (i.e., averaging the models across multiple runs) does not improve over single runs for natural language understanding  tasks \citep{pmlr-v162-wortsman22a}.
We then illustrate that sizable gains from \textit{run averaging} (\ra) are unlocked only once models are constrained a priori to converge to more structurally similar sets of parameters.
We also show that averaging the averaged checkpoints as opposed to averaging only the final models further benefits performance. 
Further, \textbf{(3)} for multilingual \fsxlt, we benchmark \ca against the established \textit{gradient surgery} method (\gs), which aims to better align gradients between languages in a batch during training for improved \fsxlt~\citep{xu-murray-2022-por}. We demonstrate that the intricate and hyperparameter-conditioned \gs performs subpar to the simple \ca. 
Finally, \textbf{(4)} we validate that benefits of \ca, \ra, and their combinations extend to a variety of experimental settings for \xlt, across a large number of different languages.




\section{Background and Related Work}
\label{sec:rw}
\sparagraph{Zero-Shot and Few-Shot \xlt}
Modern multilingual and cross-lingual NLP is underpinned by the MMTs like mBERT \citep{devlin-etal-2019-bert}, XLM(-R) \citep{lample2019cross,conneau-etal-xlmr}, or mT5 \citep{xue-etal-2021-mt5}, pretrained via language modeling (LM) objectives on web-scale corpora for 100+ languages.
The MMTs support \xlt by semantically aligning representation spaces across multiple languages. \citep{hu2020xtreme, Cao2020Multilingual}.
However, some languages `are more equal than others' in the MMTs' representation spaces~\cite{wu-dredze-2020-languages}, and the expected quality of \xlt is highly dependent on (i) the pretraining data size for the target languages, as well as on (ii) the degree of linguistic and typological (dis)similarity between the source and the target~\citep{lauscher-etal-2020-zero, ruder-etal-2021-xtreme}.

Prior work on \zsxlt thus typically aims at better aligning the language-specific subspaces for \xlt. For instance, modular approaches such as adapters~\citep{pfeiffer-etal-2020-mad, ansell-etal-2021-mad-g} and sparse subnetworks~\citep{ansell-etal-2022-composable,foroutan-etal-2022-multilingual} extend MMT to new languages by assigning a small number of language-specific parameters (i.e., modules) that can be combined with the base MMT.
Another strand of work utilizes signals from word translations or parallel data aiming to tie cross-lingual representations of languages of interest closer together~\citep{wang-etal-2019-cross,wu-dredze-2020-explicit, hu-etal-2021-explicit}.

Research on \fsxlt empirically validated that using even a handful of labeled instances in the target language along with source-language instances can considerably improve \xlt beyond \zsxlt~\citep{lauscher-etal-2020-zero, zhao-etal-2021-closer, xu-murray-2022-por, schmidt2022dontstopft}.
\fsxlt can be stabilized and improved with (i) joint training on source- and target-language data~\citep{schmidt2022dontstopft} or (ii) the so-called gradient surgery approach (\gs) which `de-conflicts' gradients between instances belonging to different languages within a training batch~\citep{xu-murray-2022-por}. 

In general, the methods that aim to boost \xlt suffer from issues such as incurring large computational costs~\cite{xu-murray-2022-por, schmidt2022dontstopft}, require additional task-annotated data~\cite{lauscher-etal-2020-zero}, and other external data (e.g., parallel data), which limits their wider portability to a multitude of possible tasks, domains, and languages \cite{ponti-etal-2019-modeling}.
\smallskip

\rparagraph{Averaging Model Weights}
As a method that is simultaneously easy to implement and inexpensive to evaluate, averaging model weights has found successful application in areas such as computer vision~\citep{huang2017snapshot,swa, pmlr-v162-wortsman22a}, machine translation~\citep{vaswani2017nips, gao-etal-2022-revisiting}, and speech processing~\citep{dong_speech,Karita2019ACS}.
The approaches can be clustered over two core axes: (i) what checkpoints to select to average model snapshots, (ii) and how to aggregate the selected model snapshots.

Stochastic weight averaging (\textsc{swa}) leverages in-training \ca to guide gradient descent towards a better generalization \citep{swa}.\footnote{However, \textsc{swa} is incompatible with adaptive optimizers and does not improve text classification over AdamW \citep{adamw}. See \url{https://github.com/timgaripov/swa/issues/6} and \url{https://discuss.huggingface.co/t/improvements-with-swa/858}.}
\ca has been proven to benefit machine translation \citep{vaswani2017nips, gao-etal-2022-revisiting}. \citet{popel_tips} recommend taking a large number of model snapshots at broad intervals. `Model souping' (\textsc{soup}) refers to averaging \textit{distinct} runs with \textit{varying hyperparameters} to further improve performance in computer vision tasks~\citep{pmlr-v162-wortsman22a}.
In monolingual NLP contexts, \citet{adamix} simultaneously train multiple adapters with \textit{consistency constraints}, allocating $2$-$10\times$ more time to their total training than what would be allocated to training only a single task adapter for GLUE tasks~\citep{wang2018glue}. In contrast, we do not expand training time or computational resources in our work. \citet{adamix} also show that subsequent adapter averaging  outperforms conventional logit ensembling.

Checkpoint selection and weighting schemes are typically devised based on validation sets \citep{pmlr-v162-wortsman22a, matena2022merging}.
One strategy is to select the $k$ checkpoints that perform best on the validation set~\citep{pmlr-v162-wortsman22a}, where $k$ is a tunable hyperparameter.
\citet{matena2022merging} show that the Fisher information matrix can be exploited to compute a weighted average of models to boost transfer across tasks.

In this work, we show that even (arguably) naive hyperparameter-free strategies to average model snapshots improve both \zsxlt and \fsxlt, and make transfer much more robust. They operate without any target-language validation data, do not increase computational demands, and even often exceed the performance of the best individual model selected using target-language validation.

\section{Methodology}
\label{sec:methodology}
Motivated by the success of weight averaging discussed in \S\ref{sec:rw}, we hypothesize that the approach might also prove effective for \xlt: weight averaging should `denoisify' idiosyncratic variation in weights of different model snapshots, which should in turn stabilize training and improve transfer.

In particular, we propose checkpoint averaging (\ca) and run averaging (\ra) of model snapshots for \zsxlt and \fsxlt.
For \textbf{\ca}, we first initialize the model with the parameters of the pretrained MMT: we refer to this set of parameters as $\theta_0$.
We then fine-tune the MMT for $T$ steps on the task data.
We store the model weights $k$ times at a regular interval of $\frac{T}{k}$ training steps.
Before inference, we then re-initialize the model with the averaged weights $\frac{1}{k} \sum_{j=1}^{k}{\theta_j} = \bar\theta$, and then use the averaged parameter set $\bar\theta$ for inference.

Run averaging (\textbf{\ra}) denotes the straightforward extension of \ca to average model snapshots taken at checkpoints across $R$ \textit{independent training runs}.
For \ra, we put forth and evaluate two different variants. 
First, we can average only the model snapshots taken at the last checkpoint of each individual run. 
The parameters at inference for this variant, termed \textbf{\ralast} are then computed as $\frac{1}{R}\sum_{i=1}^{R}\theta_{k}^{i}$. 
Here, $\theta_k^i$ denotes the final (i.e., $k$-th) model snapshot at the end of run $i$, $i=1,\ldots,R$. The second variant, termed \textbf{\raca}, combines \ca with \ra: we average all $k$ model snapshots per run over all $R$ independent runs. Effectively, we average over all $k \cdot R$ different model snapshots. The final set of model parameters used for inference is then computed as $\frac{1}{R}\sum_{j=1}^{R}\bar\theta^{i}$.


\rparagraph{Checkpoint Selection}
We only evaluate straightforward \ca and \ra strategies and dispose of more involved weighting schemes. Such schemes would require (i) either target-language validation data violating the true \xlt setup or (ii) rely on the validation data of the source language, which often yields subpar \xlt performance~\citep{keung-etal-2020-dont}.

\rparagraph{Ensuring Alignment for Run Averaging}
Prior work hinted that `plain' off-the-shelf \ra does not improve over individual models (carefully selected on validation data) on monolingual sequence classification tasks~\citep{pmlr-v162-wortsman22a}.\footnote{See Table J.1 in \citep{pmlr-v162-wortsman22a}.}
We suspect that the different random-uniform initialized classifiers from different runs draw models into unrelated training trajectories, which might also have a detrimental effect on \zsxlt.\footnote{PyTorch defaults to random-uniform initialization for linear layers \citep{7410480}.}
Pairs of random high-dimensional vectors, i.e., classifiers, are orthogonal and do not systemically align across self-contained individual runs. We have verified this hypothesis empirically in our preliminary experiments. 

Put simply, independent models converge to output representations that are orthogonal. This in turn neutralizes potential benefits of \ra, since the sets of checkpoints across runs are mutually `too distant' to complement each other.
We address this shortcoming in two steps.
We first fine-tune the model on the task in a standard fashion, yielding the first single run.
We then re-train the model $R$ times, but now we freeze all the classifiers of the $R$ models to the parameters to which the initial run converged.
This boosts alignment of the parameters of the models' respective Transformer `bodies'.
Importantly, this procedure is not required in \fsxlt, as we initialize all models with the same monolingually (source language) fine-tuned weights $\theta_{k}$, which ensures comparability across \fsxlt runs.\footnote{For \fsxlt, in our preliminary experiments we did not find variation in performance if we freeze the original classifiers stemming from monolingual English training. We observe that classifiers hardly change, as measured by the cosine similarity of classifier weights between the monolingual and multilingual checkpoints ($\geq 0.98$).}


\section{Experimental Setup}
\label{sec:experimental-setup}
\sparagraph{Tasks and Languages}
We follow prior work \cite{hu2020xtreme, lauscher-etal-2020-zero, xu-murray-2022-por, schmidt2022dontstopft} and evaluate \zsxlt and \fsxlt on benchmarks that require nuanced syntactic and semantic understanding for effective cross-lingual transfer, outlined in what follows.\footnote{Please refer to Appendix \ref{app:repro} for detailed descriptions and references of datasets by task.} We always use English as the source language.

%

\vspace{1mm}
\noindent \textit{Natural Language Inference} (NLI).
We evaluate \zsxlt on a broad range of typologically and geographically diverse NLI datasets spanning a total 37 languages: XNLI~\citep{conneau2018xnli}, IndicXNLI~\citep{https://doi.org/10.48550/arxiv.2204.08776}, JampatoisNLI~\citep{ann2022jampatois}, and AmericasNLI (AmNLI)~\cite{ebrahimi-americasnli}.
For \fsxlt experiments, we rely on 7 languages from AmericasNLI which come with sizable validation and test sets:
Aymara (\textsc{aym}), Bribri (\textsc{bzd}), Guarani (\textsc{gn}), Quechua (\textsc{quy}), Raramuri (\textsc{tar}), Shipibo-Konibo (\textsc{shp}), Wixarika (\textsc{hch}).
We feed the output \textsc{[CLS]} token of the embedded hypothesis-premise pair into the classifier.

\vspace{1mm}
\noindent \textit{Extractive QA} (TyDiQA-GoldP).
TyDiQA-GoldP consists of questions that can always be extracted from the provided gold passage \citep{clark-etal-2020-tydi}.
Our \fsxlt experiments enclose all languages: 
Arabic (\textsc{ar}),  Bengali (\textsc{bn}),  Finnish (\textsc{fi}),  Indonesian (\textsc{id}),  Korean (\textsc{ko}),  Russian (\textsc{ru}),  Swahili (\textsc{sw}), and Telegu (\textsc{te}).
The embeddings of a question-passage pair are fed into a span classifier that predicts the start and the end of the answer.

\vspace{1mm}
\noindent \textit{Named Entity Recognition} (NER).
We evaluate \xlt on a broad set of 24 languages from WikiANN~\citep{pan-etal-2017-cross} and 10 African languages from MasakhaNER~\citep{adelani-etal-2021-masakhaner}.
We choose a subset of 9 heterogeneous languages for \fsxlt: 
Arabic (\textsc{ar}), Finnish (\textsc{fi}), Hungarian (\textsc{hu}), Swahili (\textsc{sw}), Tamil (\textsc{ta}), Turkish (\textsc{tr}), Urdu (\textsc{ur}), Vietnamese (\textsc{vi}), and Chinese (\textsc{zh}).
The token representations of a sequence are fed into the classifier.

\vspace{1mm}
\noindent \textit{POS Tagging} (POS). We use the UD treebanks \citep{zeman-2020-ud} and evaluate \zsxlt on 32 languages from the XTREME benchmark~\citep{hu2020xtreme}.\footnote{We omit Kazakh, Thai, Yoruba, and Tagalog from \zsxlt results, since these languages do not comprise validation data to measure \trgdev.}
\fsxlt experiments include the following typologically diverse language sample: 
Arabic (\textsc{ar}), Basque (\textsc{eu}), Chinese (\textsc{zh}), Finnish (\textsc{fi}), German (\textsc{de}), Indonesian (\textsc{id}), Japanese (\textsc{ja}), Turkish (\textsc{tr}), and Urdu (\textsc{ur}).
The model architecture exactly matches the one used for NER.

\rparagraph{Training Setup}
XLM-R$_\text{base}$ is the main MMT in our \xlt experiments \citep{wolf-etal-2020-transformers, conneau-etal-xlmr}.\footnote{We empirically validated that our \zsxlt \& \fsxlt scores match those from other \xlt work with similar hyperparameters~\citep{wu-dredze-2020-explicit, hu-etal-2021-explicit, schmidt2022dontstopft, xu-murray-2022-por}.}$^{,}$\footnote{We preliminarily evaluated ZS-XLT experiments with XLM-V$_\text{base}$ and XLM-R$_\text{large}$, for which the results closely mimic the trends of our main results presented in Table \ref{tab:main-zsxlt-results}.}
We train models for 10 epochs with AdamW~\citep{adamw}, weight decay of $0.05$, the learning rate set to $2e^{-5}$ with a linear schedule of 10\% linear warm-up and decay, and mixed precision, unless stated otherwise.\footnote{We follow \citet{schmidt2022dontstopft} and keep hyperparameters fixed, except during ablations focusing directly on hyperparameter variation, where we analyse the impact of the number of epochs, checkpoints sampling frequency, learning rates, and scheduler.}
We simply take model snapshots at the end of each epoch.\footnote{The TyDiQA-GoldP English training portion only comprises 3,696 instances which is why we train \zsxlt models for 20 epochs. Given the size of English MNLI, we train models in \fsxlt for 1 epoch. We save snapshots at 10\% of steps in an epoch.}
The maximum input sequence length is 256 subwords for NLI, 384 with a stride of 128 for TyDiQA, and 512 for NER and POS.
We fine-tune models for \zsxlt in batches of 32 instances.
In \fsxlt experiments, we train with 4 examples per language in one batch.

\rparagraph{\fsxlt Setup}
We follow \citet{schmidt2022dontstopft} and compute a loss for examples of one language and subsequently average language-specific losses with equal weighting into a single loss.
We furthermore compare against the gradient surgery (\gs), the state-of-the-art approach for boosting multilingual \fsxlt~\citep{xu-murray-2022-por}.
For \gs, we randomly exclude one language in a batch from training. We then apply \gs for the remaining languages with respect to the held-out language.\footnote{We exclude the hyperparameter $\alpha$ denotes the share of batches that actually apply \gs from our replication of \gs, since `the values of $\alpha$ are selected empirically' \cite{xu-murray-2022-por}, which again violates the `true' \fsxlt setup.}

\rparagraph{Data Sampling and Shots}
For \fsxlt experiments, we train models with $s \in \{5, 10, 50, 100, 250\}$ target-language shots.
The training and validation splits for TyDiQA-GoldP and AmNLI are sampled from the original training and validation sets, respectively. NER and POS datasets offer sizable training portions from which we sample the `few' training shots.

\rparagraph{Random Seeds}
For \zsxlt, we initially execute 5 single runs with distinct random seeds.
We then run 5 more runs per each classifier we keep frozen from the initial runs.
For \fsxlt, we sample 5 diverse sets of $s$ shots, for each of which we conduct 5 differently seeded runs for \ra.


\rparagraph{Evaluation Metrics}
 We report average scores computed with the following metrics: accuracy for NLI, span-$F_1$ score for TyDiQA-GoldP and token-level $F_1$ for NER and POS. 
 In order to analyze robustness and sensitivity of results across different tasks and model variants, we also track and report the standard deviation over runs.

\rparagraph{Model Variants in Evaluation}
Beyond the proposed averaging strategies \ca, \raca, and \ralast (see \S\ref{sec:methodology}), we also evaluate other transfer variants outlined in what follows.
\last simply evaluates the model snapshot at the final checkpoint of a single run.
\srcdev selects the checkpoint with the corresponding model snapshot that maximizes the source-language validation metric~\citep{hu2020xtreme}.
\trgdev violates the assumption of true \xlt and assumes that the best checkpoint for \xlt can be selected using a validation set in the target language~\citep{keung-etal-2020-dont}.
This `upper-bound' single-run variant is not directly comparable to the other variants and is used for analysis purposes.\footnote{Note that, for all considered tasks and languages, the number of validation instances would always yield much more pronounced gains if used for training rather than for model selection~\citep{schmidt2022dontstopft}. Unlike other variants in our comparisons, \trgdev also requires maintaining up to $k$ models as the selected models might vary across different target languages.}

For \zsxlt, run-averaging is additionally evaluated with the `model soups' approach~\cite{pmlr-v162-wortsman22a} (termed \textsc{soup}). It comprises 5 runs spanned by varying the learning rates $\{1,2,3\}e^{{-}5}$ paired with a binary switch of using or not using a learning scheduler with 10\% warm-up.\footnote{We exclude the configuration which uses the learning rate of $3e^{{-}5}$ without a scheduler as it may diverge due to a large learning rate; this leaves the total of 6-1=5 configurations for the \textsc{soup} averaging. Corresponding single-run \zsxlt results for these configurations are in Table \ref{tab:zsxlt-ablation}.}

\section{Results and Discussion}
\label{sec:results-and-discussion}
\begin{table*}[t!]
  \begin{adjustbox}{width=\linewidth,center}
 \scriptsize
    
    \begin{tabular}{l|cgcgcgcg|cgcgcgcg}
      \toprule
      \multicolumn{1}{c}{} & \multicolumn{8}{c}{\textbf{Single Run}} & \multicolumn{8}{c}{\textbf{Ensemble}} \\
      \cmidrule(lr){2-9} \cmidrule{10-17}
      \multicolumn{1}{c}{\textbf{\zsxlt}} & \multicolumn{2}{c}{\textbf{\last}} & \multicolumn{2}{c}{\textbf{\srcdev}} & \multicolumn{2}{c}{\textbf{\trgdev}} & \multicolumn{2}{c}{\textbf{\ca}} & \multicolumn{2}{c}{\textbf{\raca}} & \multicolumn{2}{c}{\textbf{\ralast}}  & \multicolumn{2}{c}{\textbf{\soupca}} & \multicolumn{2}{c}{\textbf{\souplast}} \\
      \cmidrule(lr){2-3} \cmidrule(lr){4-5} \cmidrule(lr){6-7} \cmidrule(lr){8-9} \cmidrule(lr){10-11} \cmidrule(lr){12-13} \cmidrule(lr){14-15} \cmidrule(lr){16-17}
      Task            & \o   & $\sigma$ & \o   & $\sigma$ & \o   & $\sigma$ & \o   & $\sigma$ & \o   & $\sigma$ & \o   & $\sigma$ & \o   & $\sigma$ & \o   & $\sigma$  \\
      \hline 
      NLI             & $61.8$ & $\pm0.3$ & $61.9$ & $\pm0.3$ & $62.3$ & $\pm0.2$ & \underline{$62.8$} & $\pm0.1$ & $63.5$ & $\pm0.2$ & $63.0$ & $\pm0.3$ & \boldmath{\underline{$63.6$}} & $\pm0.4$ & $63.2$ & $\pm0.4$ \\
      TyDiQA          & $54.2$ & $\pm0.7$ & $54.8$ & $\pm1.0$ & \boldmath{\underline{$56.5$}} & $\pm0.5$ & $54.9$ & $\pm0.2$ & $54.3$ & $\pm0.5$ & $55.1$ & $\pm0.5$ & $54.3$ & $\pm0.4$ & \underline{$55.9$} & $\pm0.1$ \\
      NER             & $47.1$ & $\pm0.9$ & $47.4$ & $\pm1.1$ & \boldmath{\underline{$51.0$}} & $\pm1.4$ & $49.3$ & $\pm0.9$ & $50.0$ & $\pm0.2$ & $48.4$ & $\pm0.2$ & \underline{$50.3$} & $\pm0.4$ & $48.8$ & $\pm0.4$ \\
      POS             & $68.1$ & $\pm0.5$ & $68.1$ & $\pm0.6$ & \boldmath{\underline{$68.8$}} & $\pm0.5$ & $68.0$ & $\pm0.4$ & $68.0$ & $\pm0.4$ & \underline{$68.2$} & $\pm0.5$ & $67.8$ & $\pm0.3$ & $67.8$ & $\pm0.3$ \\
      \bottomrule
    \end{tabular}
  \end{adjustbox}
      \caption{Mean (\o) \& std. deviation ($\sigma$) of \zsxlt across 5 seeds: \last uses the final model. \srcdev (\trgdev) selects the model on a source (target) language dev set. \ca averages all checkpoints of a run. \raca (\ralast) averages all (last) checkpoints of 5 runs. \textsc{soup}s average runs with 5 sets of hyperparameters. For details, see \S\ref{sec:experimental-setup}. Best metric by group underlined, best overall metric in bold.}
  \label{tab:main-zsxlt-results}
\end{table*}

The full results for each task, dataset, and language are available in Appendix \ref{app:full-results}.
In what follows, we analyse results top-down, by type of transfer, between single runs and ensembling, along metrics, and finally datasets.  

\rparagraph{\zsxlt}
Table \ref{tab:main-zsxlt-results} summarizes the main of \zsxlt results.
We verify that our results align with relevant work for respective tasks and datasets \citep{hu-etal-2021-explicit, wu-dredze-2020-explicit}.
\smallskip

\noindent \textit{Single Run}.
%
Model snapshot selection based on the development set of the source language (\srcdev) slightly but consistently improves over the last model snapshot (\last), albeit with higher variance. \ca steadily outperforms both \last and \srcdev, and often with significantly lower variance across runs. On higher-level tasks (NLI), \ca even performs on a par with snapshot selection based on target language validation data (\trgdev), a setup that violates true \zsxlt. 
The \trgdev strategy performs best by sizable margin on POS \& NER because those test sets include a much larger number of target languages. In such a setup, \trgdev selects -- for each of the many target languages -- a snapshot tailored to a concrete language. The fact that all fair snapshot selection strategies (i.e., all except \trgdev) yield similar performance on POS suggests performance saturation when transferring from English with a single model.  



\smallskip

\noindent \textit{Ensembling}.
On tasks other than POS, ensembling (i.e., run averaging) substantially boosts \zsxlt, but only if applied with our proposed training curriculum (see ``Ensuring Alignment for Run Averaging'' in \S\ref{sec:methodology}).
The results indicate that within-run \ca is generally beneficial for ensembling too, with \{\textsc{ra, soup}\}-\textsc{ca}, in which average checkpoint-averages of individual runs, often brings gains over \{\textsc{ra, soup}\}-\textsc{last}, in which we average only the last model snapshots of each run. NER in particular seems to benefit from \ca prior to either run-averaging (\ra) or souping (i.e., averaging of runs with different hyperparameters).
\smallskip

Overall, our results indicate that \ca eliminates the need for model selection in \zsxlt. For a single run (i.e., fixed random seed) \ca clearly outperforms \srcdev -- from the \zsxlt perspective, this means that there is no need for a development set in the source language. In ensembling, \raca performs on a par with \soupca and \souplast, and better than any single run with optimal hyperparameters (cf. Table \ref{tab:zsxlt-ablation}), suggesting that it removes the need for hyperparameter optimization. 
\ca could likely be further improved by weeding out poorly performing checkpoints. This primarily facilitates \zsxlt for tasks with small training datasets, such as TyDiQA.
If target-language shots are available (cf. \fsxlt), i.e. \trgdev, models are best trained on all shots for \xlt \citep{schmidt2022dontstopft}.
\smallskip

\begin{table*}[t!]
  \begin{adjustbox}{width=\linewidth,center}
  \scriptsize
    \begin{tabular}{l|c|cgcgcgcgcg|cgcg}
      \toprule
     \multicolumn{1}{c}{\textbf{\fsxlt}} & \multicolumn{1}{c}{} & \multicolumn{10}{c}{\textbf{Single Run}} & \multicolumn{4}{c}{\textbf{Ensemble}} \\
      \cmidrule(lr){3-12} \cmidrule{13-16}
     \multicolumn{2}{c}{}    & \multicolumn{2}{c}{\textbf{\last}} & \multicolumn{2}{c}{\textbf{\textsc{gs-last}}} & \multicolumn{2}{c}{\textbf{\srcdev}} & \multicolumn{2}{c}{\textbf{\trgdev}} & \multicolumn{2}{c}{\textbf{\ca}} & \multicolumn{2}{c}{\textbf{\raca}} & \multicolumn{2}{c}{\textbf{\ralast}}                                                       \\
\cmidrule(lr){3-4} \cmidrule(lr){5-6} \cmidrule(lr){7-8} \cmidrule(lr){9-10} \cmidrule(lr){11-12} \cmidrule(lr){13-14} \cmidrule(lr){15-16} 
     Task   & Shots & \o & $\sigma$ & \o & $\sigma$ & \o & $\sigma$ & \o & $\sigma$ & \o& $\sigma$ & \o   & $\sigma$ & \o   & $\sigma$ \\
      \hline

\rule{0pt}{5.5pt}              & $5 $  & $37.0$ & $\pm1.3$ & $37.5$ & $\pm1.8$ & $36.9$ & $\pm1.3$ & \underline{$38.3$} & $\pm1.8$ & $37.6$ & $\pm1.4$ & \underline{$38.3$} & $\pm1.1$ & $38.2$ & $\pm 1.0$\\
\rule{0pt}{5.5pt}              & $10$  & $38.6$ & $\pm2.4$ & $38.5$ & $\pm3.0$ & $38.5$ & $\pm2.4$ & \boldmath{\underline{$39.7$}} & $\pm2.8$ & $39.1$ & $\pm2.7$ & \underline{$39.4$} & $\pm2.7$ & $39.1$ & $\pm 2.5$\\
\rule{0pt}{5.5pt} NLI          & $50 $ & $43.9$ & $\pm1.7$ & $43.8$ & $\pm1.7$ & $43.9$ & $\pm1.9$ & $44.3$ & $\pm1.4$ & \underline{$44.4$} & $\pm1.9$ & \boldmath{\underline{$45.0$}} & $\pm1.6$ & $44.6$ & $\pm2.1$\\
\rule{0pt}{5.5pt}              & $100$ & $45.9$ & $\pm0.3$ & $45.9$ & $\pm0.5$ & $45.9$ & $\pm0.4$ & $46.0$ & $\pm0.6$ & \underline{$46.5$} & $\pm0.5$ & \boldmath{\underline{$47.0$}} & $\pm0.8$ & $46.8$ & $\pm0.6$\\
\rule{0pt}{5.5pt}              & $250$ & $49.7$ & $\pm0.6$ & $49.7$ & $\pm0.6$ & $49.5$ & $\pm0.8$ & $49.5$ & $\pm0.7$ & \underline{$50.1$} & $\pm0.6$ & \boldmath{\underline{$50.5$}} & $\pm0.3$ & $50.4$ & $\pm0.3$\\\hline

\rule{0pt}{5.5pt}              & $5  $ & $57.9$ & $\pm0.8$ & $57.9$ & $\pm0.3$ & $57.8$ & $\pm0.9$ & \underline{$59.3$} & $\pm0.5$ & $59.0$ & $\pm0.9$ & \boldmath{\underline{$60.0$}} & $\pm0.9$ & $59.6$ & $\pm0.6$\\
\rule{0pt}{5.5pt}              & $10 $ & $60.4$ & $\pm0.8$ & $60.6$ & $\pm0.8$ & $60.0$ & $\pm0.6$ & $61.0$ & $\pm0.6$ & \underline{$61.4$} & $\pm0.8$ & \underline{$62.1$} & $\pm0.9$ & \underline{$62.1$} & $\pm0.8$\\
\rule{0pt}{5.5pt} TyDiQA       & $50 $ & $66.0$ & $\pm0.9$ & $65.9$ & $\pm1.0$ & $65.5$ & $\pm0.9$ & $66.2$ & $\pm0.7$ & \underline{$66.7$} & $\pm0.9$ & \boldmath{\underline{$67.4$}} & $\pm1.0$ & $67.0$ & $\pm0.9$\\
\rule{0pt}{5.5pt}              & $100$ & $68.2$ & $\pm0.6$ & $68.3$ & $\pm0.6$ & $68.0$ & $\pm0.6$ & $68.3$ & $\pm0.4$ & \underline{$68.9$} & $\pm0.5$ & $69.3$ & $\pm0.5$ & $69.3$ & $\pm0.4$\\
\rule{0pt}{5.5pt}              & $250$ & $71.5$ & $\pm0.5$ & $71.6$ & $\pm0.6$ & $71.2$ & $\pm0.7$ & $71.5$ & $\pm0.5$ & \underline{$72.0$} & $\pm0.5$ & \boldmath{\underline{$72.4$}} & $\pm0.5$ & $72.3$ & $\pm0.6$\\\hline

\rule{0pt}{5.5pt}              & $5  $ & $67.6$ & $\pm0.9$ & $67.1$ & $\pm1.5$ & $67.5$ & $\pm0.9$ & $68.7$ & $\pm0.9$ & \underline{$69.1$} & $\pm1.0$ & \boldmath{\underline{$70.3$}} & $\pm1.0$ & $69.7$ & $\pm1.0$ \\
\rule{0pt}{5.5pt}              & $10 $ & $70.8$ & $\pm0.9$ & $70.7$ & $\pm0.8$ & $70.8$ & $\pm0.8$ & $71.5$ & $\pm0.9$ & \underline{$72.2$} & $\pm0.8$ & \boldmath{\underline{$73.3$}} & $\pm0.9$ & $72.8$ & $\pm0.8$ \\
\rule{0pt}{5.5pt} NER          & $50 $ & $77.1$ & $\pm0.4$ & $77.1$ & $\pm0.4$ & $77.0$ & $\pm0.3$ & $77.3$ & $\pm0.3$ & \underline{$78.0$} & $\pm0.4$ & \boldmath{\underline{$78.8$}} & $\pm0.3$ & $78.6$ & $\pm0.3$ \\
\rule{0pt}{5.5pt}              & $100$ & $78.9$ & $\pm0.3$ & $78.8$ & $\pm0.2$ & $78.9$ & $\pm0.3$ & $79.0$ & $\pm0.3$ & \underline{$79.6$} & $\pm0.3$ & \boldmath{\underline{$80.2$}} & $\pm0.2$ & $80.0$ & $\pm0.3$ \\
\rule{0pt}{5.5pt}              & $250$ & $81.2$ & $\pm0.2$ & $81.2$ & $\pm0.1$ & $81.2$ & $\pm0.2$ & $81.2$ & $\pm0.2$ & \underline{$81.7$} & $\pm0.2$ & \boldmath{\underline{$82.2$}} & $\pm0.2$ & $82.1$ & $\pm0.2$ \\\hline

\rule{0pt}{5.5pt}              & $5  $ & $76.8$ & $\pm0.2$ & $76.9$ & $\pm0.4$ & $76.8$ & $\pm0.2$ & $77.1$ & $\pm0.2$ & \underline{$77.1$} & $\pm0.2$ & $77.5$ & $\pm0.2$ & \boldmath{\underline{$77.7$}} & $\pm0.2$ \\
\rule{0pt}{5.5pt}              & $10 $ & $79.2$ & $\pm0.2$ & $79.2$ & $\pm0.2$ & $79.1$ & $\pm0.2$ & $79.2$ & $\pm0.1$ & \underline{$79.4$} & $\pm0.2$ & $79.7$ & $\pm0.2$ & \boldmath{\underline{$79.9$}} & $\pm0.1$ \\
\rule{0pt}{5.5pt} POS          & $50 $ & $83.8$ & $\pm0.1$ & $83.8$ & $\pm0.1$ & $83.8$ & $\pm0.1$ & $83.8$ & $\pm0.1$ & \underline{$84.0$} & $\pm0.1$ & $84.3$ & $\pm0.1$ & \boldmath{\underline{$84.4$}} & $\pm0.1$ \\
\rule{0pt}{5.5pt}              & $100$ & $85.3$ & $\pm0.1$ & $85.4$ & $\pm0.1$ & $85.3$ & $\pm0.2$ & $85.3$ & $\pm0.2$ & \underline{$85.5$} & $\pm0.1$ & \underline{$85.8$} & $\pm0.1$ & \underline{$85.8$} & $\pm0.1$ \\
\rule{0pt}{5.5pt}              & $250$ & $86.9$ & $\pm0.1$ & $86.9$ & $\pm0.1$ & $86.9$ & $\pm0.1$ & $86.9$ & $\pm0.1$ & \underline{$87.1$} & $\pm0.1$ & \underline{$87.3$} & $\pm0.1$ & \underline{$87.3$} & $\pm0.0$ \\
      \bottomrule
    \end{tabular}
  \end{adjustbox}
      \caption{Average (\o) \& std. deviation ($\sigma$) of \fsxlt ran on 5 sets of $s$ shots for 5 seeds each: \last selects the final checkpoint. \srcdev (\trgdev) performs early stopping on a source (target) language validation set. \ca averages all checkpoints of a single run. \raca (\ralast) averages all (last) checkpoints of all runs. For details, see \S\ref{sec:experimental-setup}. Best metric by group underlined, best overall metric in bold.
    }
  \label{tab:main-fsxlt-results}
  \vspace{-0.5em}
\end{table*}

\rparagraph{\fsxlt}
Few-shot transfer results are shown in Table \ref{tab:main-fsxlt-results}.
We ensure that the results can, wherever possible, be directly compared to prior work \citep{xu-murray-2022-por,schmidt2022dontstopft}.
\smallskip

\noindent \textit{Single Run}.
Unlike in \zsxlt, \last and \srcdev result in almost identical \fsxlt performance, since they now most often select the same checkpoint.
We confirm the findings of \citet{schmidt2022dontstopft} in two regards: (1) \last gets closer to or even exceeds the oracle \trgdev as we increase the number of target-language shots; (2) using available target-language shots for training is better than leveraging them for model selection (compare, e.g., \trgdev with 50 shots against \last with 100 shots).  
Unlike in \zsxlt, in \fsxlt~\ca most often surpasses the oracle \trgdev, since all target languages (with few shots) are now part of training.
The gains over \trgdev are particularly pronounced for TyDiQA and NER and generally larger for the smaller number of shots. \ca's gains over legitimate selection strategies (\last and \srcdev) are even more pronounced. 

\smallskip

\noindent \textit{Replication of Gradient Surgery} (GS). We do not find that \gs-\last \citep{xu-murray-2022-por} improves \fsxlt, if training batches are balanced across all target languages \citep{schmidt2022dontstopft}.\footnote{\gs-\last and \gs-\srcdev yield virtually same results.}
We believe the gains that \citet{xu-murray-2022-por} report originate from the fact that, due to their small batch size ($2$-$4$), individual batches only couple English examples with those from only 1-3 target languages 
by accumulating the gradients across batches to update the model only when 32 examples are seen.\footnote{Code available at: \url{https://github.com/fe1ixxu/Mixed-Gradient-Few-Shot}}
They effectively apply \gs on many `oracle' languages instead of only one  before a parameter update (cf. Algorithm 1 of \citealp{xu-murray-2022-por}).
We thus believe that \gs mostly offsets the within-batch imbalance between languages in the original experiments.
%
Our replication further illustrates how challenging it is to reproduce the \xlt results from prior work.
Besides differing implementations, hidden effects -- such as within-batch per-language imbalance in \gs training, or other opaque hyperparameters -- hinder replication.

\smallskip

\noindent \textit{Ensembling}.
\raca and \ralast average 5 runs with different random seeds for each of five different shot setups ($\{5, ..., 250\}$).
Ensembling again brings gains, especially in configurations with smaller numbers of shots. The gains even extend to POS, a simple and saturated task on which it is otherwise difficult to improve performance. \ca is beneficial in \fsxlt ensembling too, with \raca at least matching, and often notably outperforming \ralast. Overall, the \fsxlt results corroborate the effectiveness of \ca that we noted in \zsxlt. 
\smallskip

\subsection{Further Analyses and Discussion}\label{subsec:analysis}

\begin{figure}
    \centering
    \includegraphics[scale=0.16]{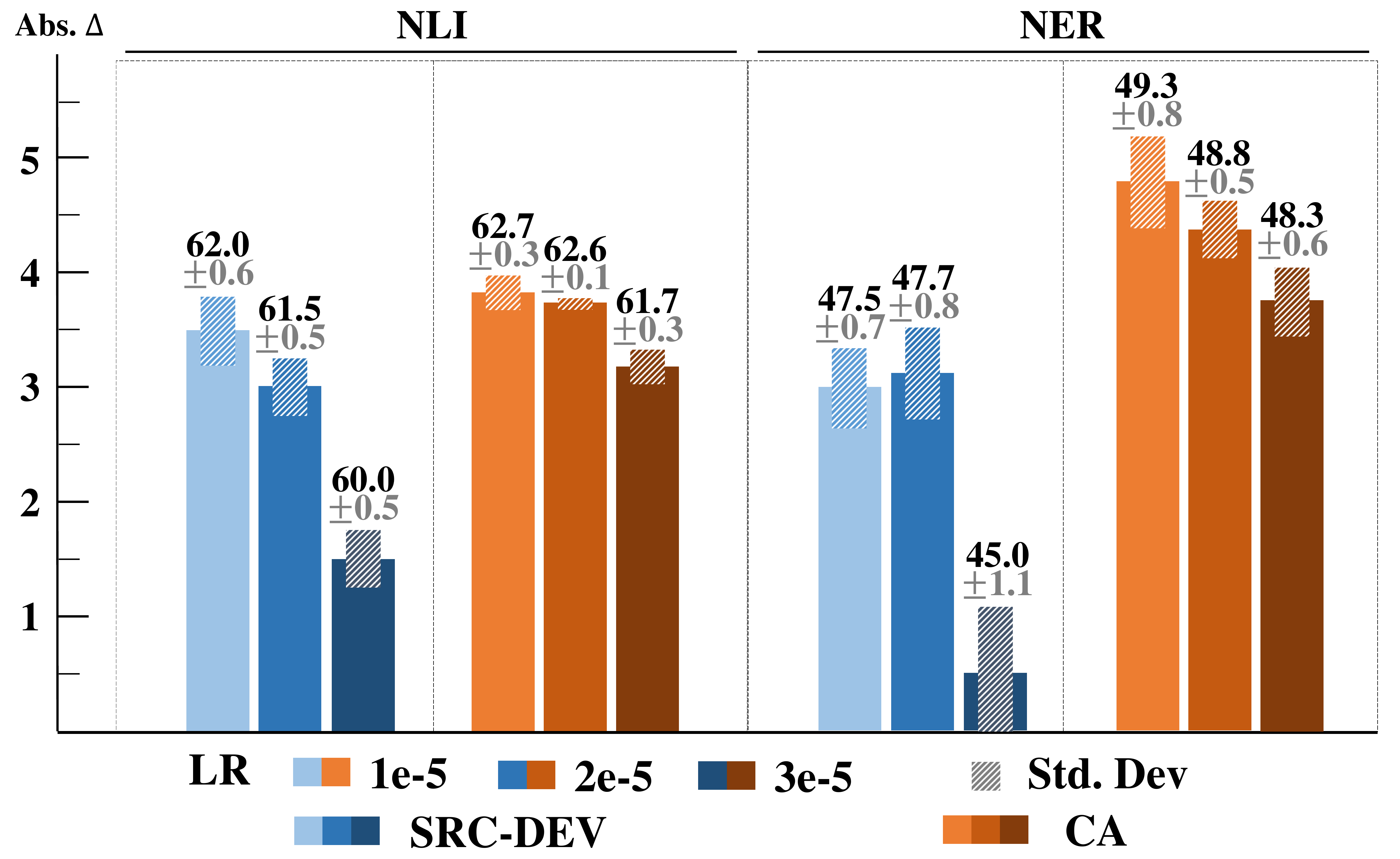}
    \caption{\zsxlt: \srcdev vs. \ca across various learning rates without a scheduler.}
    \label{fig:zsxlt-hparams}
\vspace{-0.5em}
\end{figure}

\begin{table*}[t!]
 \footnotesize
  \begin{adjustbox}{width=\linewidth,center}
  \setlength{\tabcolsep}{1.7pt}
    \begin{tabular}{c|c|cgcgcgcg|cgcgcgcg|cgcgcgcg|cgcgcgcg}
      \toprule
\multicolumn{1}{c}{} & \multicolumn{1}{c}{} & \multicolumn{8}{c}{\textbf{NLI}} & \multicolumn{8}{c}{\textbf{TyDiQA}} & \multicolumn{8}{c}{\textbf{NER}} & \multicolumn{8}{c}{\textbf{POS}} \\
\cmidrule(lr){3-10}  \cmidrule(lr){11-18} \cmidrule(lr){19-26} \cmidrule(lr){27-34}
\multicolumn{1}{c}{} & \multicolumn{1}{c}{} & \multicolumn{2}{c}{\textbf{\last}} & \multicolumn{2}{c}{\textbf{\sdev}} & \multicolumn{2}{c}{\textbf{\tdev}} & \multicolumn{2}{c}{\textbf{\ca}} &\multicolumn{2}{c}{\textbf{\last}} & \multicolumn{2}{c}{\textbf{\sdev}} & \multicolumn{2}{c}{\textbf{\tdev}} & \multicolumn{2}{c}{\textbf{\ca}} &\multicolumn{2}{c}{\textbf{\last}} & \multicolumn{2}{c}{\textbf{\sdev}} & \multicolumn{2}{c}{\textbf{\tdev}} & \multicolumn{2}{c}{\textbf{\ca}} &\multicolumn{2}{c}{\textbf{\last}} & \multicolumn{2}{c}{\textbf{\sdev}} & \multicolumn{2}{c}{\textbf{\tdev}} & \multicolumn{2}{c}{\textbf{\ca}} \\
\cmidrule(lr){3-4} \cmidrule(lr){5-6} \cmidrule(lr){7-8} \cmidrule(lr){9-10}  \cmidrule(lr){11-12} \cmidrule(lr){13-14} \cmidrule(lr){15-16} \cmidrule(lr){17-18} \cmidrule(lr){19-20} \cmidrule(lr){21-22} \cmidrule(lr){23-24} \cmidrule(lr){25-26} \cmidrule(lr){27-28} \cmidrule(lr){29-30} \cmidrule(lr){31-32} \cmidrule(lr){33-34} 

S & B & \o   & $\sigma$ & \o   & $\sigma$ & \o   & $\sigma$ & \o   & $\sigma$ & \o   & $\sigma$ & \o   & $\sigma$ & \o   & $\sigma$ & \o   & $\sigma$ & \o   & $\sigma$ & \o   & $\sigma$ & \o   & $\sigma$ & \o   & $\sigma$ & \o   & $\sigma$ & \o   & $\sigma$ & \o   & $\sigma$ & \o   & $\sigma$ \\\hline
\multirow{3}{*}{{0}}    &  \textonehalf  & $62.1$ & $0.2$ & $62.3$ & $0.2$ & $62.6$ & $0.2$ & $62.8$ & $0.1$ & $54.4$ & $1.3$ & $54.0$ & $1.2$ & $55.4$ & $0.6$ & $52.7$ & $0.7$ & $48.8$ & $0.5$ & $48.8$ & $0.5$ & $50.4$ & $1.1$ & $49.3$ & $0.9$ & $67.9$ & $0.3$ & $67.9$ & $0.3$ & $68.2$ & $0.3$ & $67.7$ & $0.3$\\
                        & $1$ & $61.8$ & $0.3$ & $61.9$ & $0.3$ & $62.3$ & $0.2$ & $62.8$ & $0.1$ & $54.2$ & $0.7$ & $54.8$ & $1.1$ & $56.5$ & $0.5$ & $54.9$ & $0.2$ & $47.1$ & $0.9$ & $47.4$ & $1.1$ & $51.0$ & $1.4$ & $49.3$ & $0.9$ & $68.1$ & $0.5$ & $68.1$ & $0.6$ & $68.8$ & $0.5$ & $68.0$ & $0.4$\\ 
                        & $2$ & $61.3$ & $0.2$ & $61.2$ & $0.2$ & $62.3$ & $0.1$ & $62.4$ & $0.2$ & $54.8$ & $0.4$ & $54.6$ & $0.8$ & $56.5$ & $0.5$ & $55.0$ & $0.6$ & $47.0$ & $0.7$ & $46.9$ & $0.7$ & $51.5$ & $0.6$ & $49.1$ & $0.6$ & $67.8$ & $0.5$ & $67.9$ & $0.4$ & $69.3$ & $0.3$ & $68.1$ & $0.4$\\\hline\hline
\multirow{3}{*}{{10}}   & \textonehalf  & $38.4$ & $2.3$ & $38.5$ & $2.3$ & $38.9$ & $2.7$ & $38.8$ & $2.5$ & $60.1$ & $0.4$ & $59.8$ & $0.4$ & $60.4$ & $0.3$ & $60.7$ & $0.6$ & $71.3$ & $1.0$ & $71.3$ & $1.0$ & $71.7$ & $0.9$ & $72.1$ & $0.8$ & $79.0$ & $0.2$ & $79.0$ & $0.2$ & $79.0$ & $0.2$ & $79.1$ & $0.3$\\
                        & $1$ & $38.6$ & $2.4$ & $38.5$ & $2.4$ & $39.7$ & $2.8$ & $39.1$ & $2.7$ & $60.4$ & $0.8$ & $60.0$ & $0.6$ & $61.0$ & $0.6$ & $61.4$ & $0.8$ & $70.8$ & $0.9$ & $70.8$ & $0.8$ & $71.5$ & $0.9$ & $72.2$ & $0.8$ & $79.2$ & $0.2$ & $79.1$ & $0.1$ & $79.2$ & $0.1$ & $79.4$ & $0.2$\\
                        & $2$ & $38.7$ & $2.6$ & $38.9$ & $2.9$ & $39.6$ & $2.7$ & $39.3$ & $3.0$ & $60.8$ & $0.8$ & $60.2$ & $1.0$ & $61.6$ & $0.7$ & $62.2$ & $0.7$ & $70.5$ & $0.9$ & $70.4$ & $0.9$ & $71.7$ & $1.0$ & $72.2$ & $0.8$ & $79.1$ & $0.2$ & $79.1$ & $0.2$ & $79.4$ & $0.1$ & $79.6$ & $0.1$\\\hline
\multirow{3}{*}{{250}}  & \textonehalf  & $49.9$ & $0.7$ & $49.9$ & $0.6$ & $49.5$ & $0.7$ & $50.1$ & $0.8$ & $71.6$ & $0.4$ & $71.1$ & $0.4$ & $71.3$ & $0.4$ & $71.7$ & $0.5$ & $81.2$ & $0.1$ & $81.1$ & $0.2$ & $81.3$ & $0.2$ & $81.7$ & $0.1$ & $86.9$ & $0.1$ & $86.9$ & $0.1$ & $86.9$ & $0.1$ & $87.0$ & $0.1$\\
                        & $1$ & $49.7$ & $0.6$ & $49.5$ & $0.8$ & $49.5$ & $0.7$ & $50.1$ & $0.6$ & $71.5$ & $0.5$ & $71.2$ & $0.7$ & $71.5$ & $0.5$ & $72.0$ & $0.1$ & $81.2$ & $0.2$ & $81.2$ & $0.2$ & $81.2$ & $0.2$ & $81.7$ & $0.1$ & $86.9$ & $0.1$ & $86.9$ & $0.1$ & $86.9$ & $0.1$ & $87.1$ & $0.1$\\
                        & $2$ & $50.0$ & $0.7$ & $49.1$ & $0.6$ & $49.7$ & $0.8$ & $50.5$ & $0.8$ & $71.7$ & $0.6$ & $71.3$ & $0.5$ & $71.8$ & $0.3$ & $72.6$ & $0.5$ & $81.1$ & $0.2$ & $81.1$ & $0.2$ & $81.2$ & $0.2$ & $81.9$ & $0.1$ & $86.8$ & $0.1$ & $86.8$ & $0.1$ & $86.8$ & $0.1$ & $87.1$ & $0.1$\\
      \bottomrule
    \end{tabular}
  \end{adjustbox}
  \caption{Ablation of budget (B) on \xlt:  \textonehalf\ (2) B perform half (double) the steps and half (double) the checkpoints of 1 B. \zsxlt \& \fsxlt experiments are not comparable. \sdev = \srcdev, \tdev = \trgdev. }
  \label{tab:xlt-epoch-ablation}
  \vspace{-0.5em}
\end{table*}

\begin{table}[t!]
 \begin{adjustbox}{width=\linewidth,center}
 \footnotesize
    \begin{tabular}{l|c|cc|cc}
      \toprule
          \multicolumn{2}{c}{}  & \multicolumn{2}{c}{\textbf{Single Run}} & \multicolumn{2}{c}{\textbf{Ensemble}} \\\hline
          \multicolumn{2}{c}{}  & \textbf{\last} & \textbf{\ca} & \textbf{\raca} & \textbf{\ralast} \\\hline
      Task            & Shots & \o   & \o    & \o    & \o   \\
      \hline 
 \multirow{5}{*}{{NLI}}     & $5  $ & $61.4$ & $62.2$ & $62.9$ & $62.7$ \\
                            & $10 $ & $61.7$ & $62.5$ & $63.2$ & $62.9$ \\
                            & $50 $ & $62.6$ & $63.3$ & $64.0$ & $63.8$ \\
                            & $100$ & $62.9$ & $63.6$ & $64.3$ & $64.1$ \\
                            & $250$ & $63.1$ & $63.7$ & $64.4$ & $64.1$ \\\hline
 \multirow{5}{*}{{NER}}     & $5  $ & $21.8$ & $23.6$ & $24.1$ & $23.0$ \\
                            & $10 $ & $23.2$ & $25.0$ & $25.9$ & $24.5$ \\
                            & $50 $ & $26.2$ & $28.4$ & $29.1$ & $27.5$ \\
                            & $100$ & $27.7$ & $29.5$ & $30.1$ & $29.0$ \\
                            & $250$ & $29.9$ & $32.1$ & $33.0$ & $31.4$ \\
      \bottomrule
    \end{tabular}
  \end{adjustbox}
      \caption{\zsxlt with multilingual models of Table \ref{tab:main-fsxlt-results}.}
  \label{tab:fsxlt-multilingual-zsxlt}
\vspace{-0.5em}
\end{table}


To test the robustness of \ca, we run additional ablations: we compare \zsxlt results for models trained (1) with different learning rates; and (2) 
under different computational budgets.
\smallskip

\rparagraph{Hyperparameters for \zsxlt}
We repeat \zsxlt experiments with LRs of $\{1,2,3\}e^{{-}5}$, with and without a scheduler of 10\% warm-up and subsequent decay (5 runs for each combination).
Figure \ref{fig:zsxlt-hparams} summarizes the findings for \srcdev and \ca on NLI and NER (complete results are in Table \ref{tab:zsxlt-ablation} in the Appendix).
In comparison with \srcdev, \ca reduces the variance in results between runs with different learning rates as well within different runs with the same learning rate for both tasks.  
This yields further benefits.
\ca, unlike \srcdev, allows for \zsxlt performance to depend much less on the selection of learning rates, rendering hyperparameter tuning less important for the final performance.  
This also in part explains why \raca further improves over \ralast: it averages more robust models from individual runs (cf. `\textsc{soup}s` in Table \ref{tab:main-zsxlt-results}).
%
%
This ablation contributes to the explanation of why \zsxlt results greatly differ in the literature \citep{keung-etal-2020-dont}. For example, with learning rate scheduling, \last deteriorates much more severely than \srcdev (especially at higher learning rates). This again stresses the need for strategies such as \ca that stabilize \xlt performance across runs and hyperparameters.  

\rparagraph{Training Duration for \xlt}
Table \ref{tab:xlt-epoch-ablation} presents experiments for \zsxlt and \fsxlt with $\{10, 250\}$ shots, in which we halve and double the number of training steps.\footnote{For \zsxlt in TyDiQA-GoldP, we increase the number of epochs from 20 to 30.}
In \zsxlt, the takeaways align with the original experiments of Table \ref{tab:main-zsxlt-results}.
For \fsxlt, \ca gains further ground relative to \last and \srcdev in prolonged training.
This particularly proves true when only 10 shots per target language are available.
Performance may be further improved by distributing the added compute budget more diversely.
Rather than doubling the steps along a single trajectory that well converges in the original compute budget (i.e., 1 B), averaging two runs likely mitigates unfavorable variation within the snapshots of each run.
Our \ra-variants in the main \fsxlt results in Table \ref{tab:main-fsxlt-results} hint at that this likely proves true in \fsxlt as averaging across runs consistently yielded sizable improvements.
We however leave such experiments to future work.

\rparagraph{\zsxlt for Multilingual Models}
We additionally test the behaviour of multilingual models -- trained on large source-language dataset and a multilingual dataset consisting of few-shots of target languages (included in \fsxlt training) -- in \zsxlt to few remaining unseen languages: (1) for NLI -- 3 languages from AmNLI~\citep{ebrahimi-americasnli}, all languages from JampatoisNLI~\citep{ann2022jampatois} and IndicXNLI~\citep{https://doi.org/10.48550/arxiv.2204.08776}; (2) for NER, all languages from  MasakhaNER~\citep{adelani-etal-2021-masakhaner}.
Table \ref{tab:fsxlt-multilingual-zsxlt} summarizes the results of this experiment. We again observe similar trends. Within a single run, \ca yields large gains, now even more pronounced with more multilingual shots. \raca continues to generally outperform \ralast in the ensembling setup.  
Interestingly, for NER, single-run \ca even outperforms the \ralast ensemble.
Results of this realistic transfer of a multilingually trained model to a new (unseen) language confirms the utility of model averaging in \xlt.

\section{Conclusion}
\label{sec:conclusion}
It is hard to meaningfully compare prior work on \xlt: experimental setups are opaque and models are (often unreportedly) selected based on performance on English development data or even target-language instances. On the one hand, selecting models based on target-language performance violates the `zero-shot' assumption of \zsxlt and overestimates performance in both \zsxlt and \fsxlt.
Model selection on source-language data, on the other hand, has been proven unreliable \cite{keung-etal-2020-dont}.
Further, reproducing existing work on \xlt is unwieldy: even if code and models are available, replication incurs a significant overhead in terms of integration efforts and computing resources.
In this work, we propose to \textit{average checkpoints} (\ca) stored periodically in training as a simple, computationally cheap, and effective baseline for \xlt that remedies for all of the above.
We show that (1) \ca consistently improves both \zsxlt and \fsxlt over model selection based on source-language data \xlt baselines and (2) brings stability in performance across different runs.
Further, we propose a curriculum training that involves freezing of classifier's parameters, allowing \ca benefits to propagate to ensembling, i.e., averaging of models from independent runs.
We hope that future works adopts \ca as a competitive and robust baseline. This would lead to more transparency and fairness in \xlt evaluation, leading to more trustworthy results. 

\section*{Limitations}

The primary weakness of `fairly' averaging model weights for \xlt is that \textit{sensible} checkpoints need to be averaged.
This manifests, for instance, in hyperparameter ablation for \zsxlt on TyDiQA-GoldP.
TyDiQA-GoldP is a complex task with merely 3,696 training instances that observes unusual training dynamics.
On such a dataset, the early checkpoints often underperform models that (nearly) have converged, especially if training utilizes low learning rates with schedulers.
Here, \srcdev could be used to weed out underperforming checkpoints, such that \ca then \textit{always} exceeds the baseline that performs model selection on source-language validation data.
Whenever the English training portion is sizable -- like in our other tasks -- checkpoint averaging is consistently beneficial.
Our experiments also demonstrate that \xlt behaves differently by task.
Averaging checkpoints consequently might affect other tasks differently like, for instance, document classification that reason about long contexts or retrieval tasks like Tatoeba that jointly require sequence- and word-level semantics.
Another dimension we did not explore further due to a limited compute budget is how to ensure \textit{best} that monolingual models are aligned for run averaging.
For instance, it may not be required or even desirable to keep classifiers frozen throughout the second step of our proposed training curriculum (\S\ref{sec:methodology}), as we would ideally also want to average out idiosyncratic noise of the original classifier.

\section*{Acknowledgments}
We thank the state of Baden-Württemberg for its support through access to the bwHPC. Ivan Vuli\'{c} is supported by a personal Royal Society University Research Fellowship \textit{`Inclusive and Sustainable Language Technology for a Truly Multilingual World'} (no 221137; 2022--).

\bibliography{anthology,custom}
\bibliographystyle{acl_natbib}

\appendix

\section{Appendix}
\label{sec:appendix}

%

\subsection{Reproduction Details}
\label{app:repro}

\noindent \textbf{Code}. Our code is available at: \url{https://github.com/fdschmidt93/free-lunch-xlt}

\smallskip

\noindent \textbf{Model architectures.} All models rely on the \texttt{AutoModelFor\{SequenceClassification, TokenClassification, QuestionAnswering\}} of \texttt{xlm-roberta-base} implementations fitting the corresponding task of the \texttt{transformers} library \citep{wolf-etal-2020-transformers}.

\smallskip

\noindent \textbf{Compute Requirements.} All the experiments were run on a single V100 with 32GB VRAM.
The total required GPU time (training \& evaluation) per run for \zsxlt is c.2.75 hours and \fsxlt 5 hours on average.
We repeated each set of experiments at least 5 (and up to 25) times to reliably measure mean and standard deviation of performance.
For \zsxlt, we trained, per task, 5 initial models, 25 $\times$ 2 additional models to evaluate \ra and \textsc{soup}s (i.e. 5 varying classification heads, cf \S\ref{sec:methodology}), and 20 further models per configuration for each hyperparameter ablation.
We trained 25 models per $s$ shots in \fsxlt (i.e. 5 sets of different $s$ shots with 5 runs each).
We roughly estimate that total GPU time accumulates to 6,400 hours across all experiments.

\smallskip

\noindent \textbf{Further Dataset Details}. All datasets are accessed via the \texttt{datasets} library \citep{lhoest-etal-2021-datasets}. We sub-sample shots for datasets that do not comprise a training split for \fsxlt experiments as follows. We first randomly shuffle the validation split with one of seed $s \in \{42, \dots, 46\}$ with the built-in \texttt{datasets} \texttt{shuffle} method and then gather the initial $\{5, 10, 50, 100, 250\}$ instances as training shots for our \xlt experiments. We then validate our models on the the $|N_{D}| - 500$ remaining instances to measure \trgdev performance.

%

\smallskip

\noindent \textit{Natural Language Inference} (NLI). As is custom, we use the sizable training split of \href{https://huggingface.co/datasets/glue}{MNLI} \citep{N18-1101} as our high-resource training dataset with 393K training instances for English.  The source-language validation split is the development portion of \href{https://huggingface.co/datasets/xnli}{XNLI} \cite{conneau2018xnli}. We furthermore evaluate on \href{https://huggingface.co/datasets/Divyanshu/indicxnli}{IndicXNLI}~\citep{https://doi.org/10.48550/arxiv.2204.08776}, \href{https://github.com/fdschmidt93/free-lunch-xlt/blob/master/trident-xtreme/jampatois_dataset.py}{JampatoisNLI}~\citep{ann2022jampatois}, and \href{https://huggingface.co/datasets/americas_nli}{AmericasNLI} (AmNLI)~\cite{ebrahimi-americasnli}.

\smallskip

\noindent \textit{Extractive QA} (TyDiQA-GoldP). For \href{https://github.com/fdschmidt93/free-lunch-xlt/blob/master/trident-xtreme/tydiqa_goldp_dataset.py}{TyDiQA-GoldP}, we sub-sample training and validation instances as per the procedure noted above from all the training sets and use the official validation splits for testing \citep{clark-etal-2020-tydi}.
We compute \srcdev on the bases of the 440 `test' set instances of English, as the training split merely comprises 3,696 instances. This favors \srcdev compared to other selection strategies based on the source language, as another 10\% of the training data are used for early stopping.

\smallskip

\noindent \textit{Named Entity Recognition} (NER). As with other tasks, we access both \href{https://huggingface.co/datasets/wikiann}{WikiANN} and \href{https://huggingface.co/datasets/masakhaner}{MasakhaNER} via the Huggingface \href{https://github.com/huggingface/datasets}{\texttt{datasets}} library \citep{lhoest-etal-2021-datasets}. We train monolingual models for \zsxlt on the English training portion of Wikiann.

\smallskip

\noindent \textit{POS Tagging} (POS). We use the UD treebanks \citep{zeman-2020-ud} and evaluate \zsxlt on 32 languages from the \href{https://huggingface.co/datasets/xtreme}{XTREME} benchmark~\citep{hu2020xtreme}. We omit Kazakh, Thai, Yoruba, and Tagalog from \zsxlt results, since these languages do not comprise validation data to measure \trgdev.

\smallskip

\noindent \textbf{Sample Implementation.}
The below exemplary code is a simple implementation to average the \texttt{state\_dict} of identical PyTorch models.
The resulting averaged parameter can the been used to reinitialize the model with \texttt{model.load\_state\_dict(state\_dict)}.

\begin{minted}[
fontsize=\footnotesize,
]{python}
import torch

def average_weights(
    state_dicts: list[dict[str, torch.Tensor]]
) -> dict[str, torch.Tensor]:
    """Avg. state_dicts of models
       with same architecture."""
    avg_state_dict = {}
    K = len(state_dicts)
    for (
        name,
        params,
    ) in avg_state_dict.items():
        if params.is_floating_point():
            avg_state_dict[name] = params / K
    for state_dict in state_dicts[1:]:
        for (
            name,
            params,
        ) in avg_state_dict.items():
            if params.is_floating_point():
                avg_state_dict[name] += (
                    state_dict[name] / K
                )
    return avg_state_dict
\end{minted}


\begin{table*}[t]
\subsection{Full Results}\label{app:full-results}
 \footnotesize
  \begin{adjustbox}{width=\linewidth,center}
\begin{tabular}{c|c|cgcgcgcgcgcgcgcg}
      \toprule
      \multicolumn{2}{c}{} & \multicolumn{4}{c}{\textbf{\last}} & \multicolumn{4}{c}{\textbf{\srcdev}} & \multicolumn{4}{c}{\textbf{\trgdev}}   & \multicolumn{4}{c}{\textbf{\ca}}                                                             \\
      \cmidrule(lr){3-6} \cmidrule(lr){7-10} \cmidrule(lr){11-14} \cmidrule(lr){15-18} 
      \multicolumn{2}{c}{Scheduler} &\multicolumn{2}{c}{None} & \multicolumn{2}{c}{$10\%$} &\multicolumn{2}{c}{None} & \multicolumn{2}{c}{$10\%$} &\multicolumn{2}{c}{None} & \multicolumn{2}{c}{$10\%$} &\multicolumn{2}{c}{None} & \multicolumn{2}{c}{$10\%$}  \\
      \cmidrule(lr){3-4} \cmidrule(lr){5-6} \cmidrule(lr){7-8} \cmidrule(lr){9-10} \cmidrule(lr){11-12} \cmidrule(lr){13-14} \cmidrule(lr){15-16} \cmidrule(lr){17-18}
      
{} & LR & \o & $\sigma$ & \o & $\sigma$ & \o & $\sigma$ & \o & $\sigma$ & \o & $\sigma$ & \o & $\sigma$ & \o & $\sigma$ & \o & $\sigma$ \\\hline
\multirow{3}{*}{{\rotatebox[origin=c]{90}{NLI}}}   & $1e{-}5$   & 61.0 & $\pm$0.2  & 61.9 & $\pm$0.2  & 62.0 & $\pm$0.6  & 62.4 & $\pm$0.4  & 62.7 & $\pm$0.2  & 63.1 & $\pm$0.6  & 62.7 & $\pm$0.3 & 62.7 & $\pm$0.1\\
& $2e{-}5$                                                      & 59.6 & $\pm$0.4  & 61.8 & $\pm$0.3  & 61.5 & $\pm$0.5  & 61.9 & $\pm$0.3  & 62.6 & $\pm$0.4  & 62.5 & $\pm$0.2  & 62.6 & $\pm$0.1 & 62.8 & $\pm$0.1 \\
& $3e{-}5$                                                      & 57.3 & $\pm$0.3  & 61.1 & $\pm$0.3  & 60.0 & $\pm$0.5  & 61.2 & $\pm$0.4  & 61.7 & $\pm$0.8  & 62.4 & $\pm$0.2  & 61.7 & $\pm$0.3 & 62.7 & $\pm$0.3 \\\hline
\multirow{3}{*}{{\rotatebox[origin=c]{90}{QA}}} & $1e{-}5$      & 52.8 & $\pm$1.1  & 51.9 & $\pm$1.0  & 52.2 & $\pm$1.2  & 52.4 & $\pm$1.0  & 54.2 & $\pm$0.7  & 53.9 & $\pm$0.7  & 52.4 & $\pm$1.4 & 50.9 & $\pm$0.4 \\
\bottomrule
    \end{tabular}
  \end{adjustbox}
  \caption{Ablation of hyperparameters on \zsxlt: \last selects the final checkpoint. \srcdev (\trgdev) performs early stopping on a source (target) language validation set. \ca averages all checkpoints of a single run.}
  \label{tab:zsxlt-ablation}
\end{table*}

\begin{table*}[t]
\subsubsection{\zsxlt Results}
 \footnotesize
  \begin{adjustbox}{width=\linewidth,center}
\begin{tabular}{l|cgcgcgcgcgcgcgcgcgcgcgcgcgcg|cg}
\toprule
\textbf{Languages} & \multicolumn{2}{c}{\textbf{\textsc{ar}}} & \multicolumn{2}{c}{\textbf{\textsc{bg}}} & \multicolumn{2}{c}{\textbf{\textsc{de}}} & \multicolumn{2}{c}{\textbf{\textsc{el}}} & \multicolumn{2}{c}{\textbf{\textsc{es}}} & \multicolumn{2}{c}{\textbf{\textsc{fr}}} & \multicolumn{2}{c}{\textbf{\textsc{hi}}} & \multicolumn{2}{c}{\textbf{\textsc{ru}}} & \multicolumn{2}{c}{\textbf{\textsc{sw}}} & \multicolumn{2}{c}{\textbf{\textsc{th}}} & \multicolumn{2}{c}{\textbf{\textsc{tr}}} & \multicolumn{2}{c}{\textbf{\textsc{ur}}} & \multicolumn{2}{c}{\textbf{\textsc{vi}}} & \multicolumn{2}{c}{\textbf{\textsc{zh}}} & \multicolumn{2}{c}{\textbf{\textsc{Total}}} \\
            Metric &                   \o & $\sigma$ &                   \o & $\sigma$ &                   \o & $\sigma$ &                   \o & $\sigma$ &                   \o & $\sigma$ &                   \o & $\sigma$ &                   \o & $\sigma$ &                   \o & $\sigma$ &                   \o & $\sigma$ &                   \o & $\sigma$ &                   \o & $\sigma$ &                   \o & $\sigma$ &                   \o & $\sigma$ &                   \o & $\sigma$ &                      \o & $\sigma$ \\
\midrule
             \last &                 71.1 &      0.5 &                 76.8 &      0.4 &                 75.2 &      0.5 &                 74.8 &      0.6 &                 77.9 &      0.3 &                 77.2 &      0.4 &                 68.9 &      0.6 &                 74.9 &      0.3 &                 61.7 &      0.4 &                 70.8 &      0.5 &                 71.2 &      0.2 &                 64.1 &      0.6 &                 73.7 &      0.5 &                 73.4 &      0.4 &                    72.3 &      0.2 \\
\bottomrule
\end{tabular}
  \end{adjustbox}
  \caption{\zsxlt to XNLI~\citep{conneau2018xnli}.}
\end{table*}

\begin{table*}[t]
 \footnotesize
  \begin{adjustbox}{width=\linewidth,center}
\begin{tabular}{l|cgcgcgcgcgcgcgcgcgcg|cg}
\toprule
\textbf{Languages} & \multicolumn{2}{c}{\textbf{\textsc{aym}}} & \multicolumn{2}{c}{\textbf{\textsc{bzd}}} & \multicolumn{2}{c}{\textbf{\textsc{gn}}} & \multicolumn{2}{c}{\textbf{\textsc{hch}}} & \multicolumn{2}{c}{\textbf{\textsc{quy}}} & \multicolumn{2}{c}{\textbf{\textsc{shp}}} & \multicolumn{2}{c}{\textbf{\textsc{tar}}} & \multicolumn{2}{c}{\textbf{\textsc{cni}}} & \multicolumn{2}{c}{\textbf{\textsc{nah}}} & \multicolumn{2}{c}{\textbf{\textsc{oto}}} & \multicolumn{2}{c}{\textbf{\textsc{Total}}} \\
            Metric &                    \o & $\sigma$ &                    \o & $\sigma$ &                   \o & $\sigma$ &                    \o & $\sigma$ &                    \o & $\sigma$ &                    \o & $\sigma$ &                    \o & $\sigma$ &                    \o & $\sigma$ &                    \o & $\sigma$ &                    \o & $\sigma$ &                      \o & $\sigma$ \\
\midrule
             \last &                  38.7 &      2.1 &                  40.1 &      1.3 &                 40.3 &      1.2 &                  38.2 &      1.0 &                  38.1 &      1.2 &                  40.4 &      1.0 &                  37.9 &      1.5 &                  39.7 &      1.4 &                  42.7 &      1.4 &                  39.7 &      1.7 &                    39.6 &      0.7 \\
           \srcdev &                  38.6 &      2.0 &                  40.3 &      1.5 &                 40.4 &      1.4 &                  37.8 &      0.8 &                  38.7 &      1.3 &                  40.4 &      1.0 &                  38.2 &      1.3 &                  39.8 &      1.1 &                  42.6 &      1.2 &                  39.4 &      1.6 &                    39.6 &      0.7 \\
\bottomrule
\end{tabular}
  \end{adjustbox}
  \caption{\zsxlt to AmNLI~\citep{ebrahimi-americasnli}.}
\end{table*}

\begin{table*}[t]
 \footnotesize
  \begin{adjustbox}{width=\linewidth,center}
\begin{tabular}{l|cgcgcgcgcgcgcgcgcgcgcg|cg}
\toprule
\textbf{Languages} & \multicolumn{2}{c}{\textbf{\textsc{as}}} & \multicolumn{2}{c}{\textbf{\textsc{bn}}} & \multicolumn{2}{c}{\textbf{\textsc{gu}}} & \multicolumn{2}{c}{\textbf{\textsc{hi}}} & \multicolumn{2}{c}{\textbf{\textsc{kn}}} & \multicolumn{2}{c}{\textbf{\textsc{ml}}} & \multicolumn{2}{c}{\textbf{\textsc{mr}}} & \multicolumn{2}{c}{\textbf{\textsc{or}}} & \multicolumn{2}{c}{\textbf{\textsc{pa}}} & \multicolumn{2}{c}{\textbf{\textsc{ta}}} & \multicolumn{2}{c}{\textbf{\textsc{te}}} & \multicolumn{2}{c}{\textbf{\textsc{Total}}} \\
            Metric &                   \o & $\sigma$ &                   \o & $\sigma$ &                   \o & $\sigma$ &                   \o & $\sigma$ &                   \o & $\sigma$ &                   \o & $\sigma$ &                   \o & $\sigma$ &                   \o & $\sigma$ &                   \o & $\sigma$ &                   \o & $\sigma$ &                   \o & $\sigma$ &                      \o & $\sigma$ \\
\midrule
             \last &                 61.8 &      0.5 &                 69.3 &      0.3 &                 69.3 &      0.5 &                 73.3 &      0.3 &                 70.2 &      0.6 &                 70.1 &      0.3 &                 68.1 &      0.3 &                 67.6 &      0.5 &                 68.6 &      0.4 &                 69.2 &      0.4 &                 68.5 &      0.6 &                    68.7 &      0.2 \\
           \srcdev &                 61.8 &      0.6 &                 69.6 &      0.4 &                 69.6 &      0.6 &                 73.5 &      0.5 &                 70.3 &      0.5 &                 70.2 &      0.4 &                 68.3 &      0.5 &                 67.8 &      0.4 &                 69.0 &      0.6 &                 69.4 &      0.5 &                 68.9 &      0.7 &                    69.0 &      0.3 \\
\bottomrule
\end{tabular}
  \end{adjustbox}
  \caption{\zsxlt to IndicXNLI~\citep{https://doi.org/10.48550/arxiv.2204.08776}.}
\end{table*}

\begin{table*}[t]
 \footnotesize
  \begin{adjustbox}{width=\linewidth,center}
\begin{tabular}{l|cgcgcgcgcgcgcgcg|cg}
\toprule
\textbf{Languages} & \multicolumn{2}{c}{\textbf{\textsc{ar}}} & \multicolumn{2}{c}{\textbf{\textsc{bn}}} & \multicolumn{2}{c}{\textbf{\textsc{fi}}} & \multicolumn{2}{c}{\textbf{\textsc{id}}} & \multicolumn{2}{c}{\textbf{\textsc{ko}}} & \multicolumn{2}{c}{\textbf{\textsc{ru}}} & \multicolumn{2}{c}{\textbf{\textsc{sw}}} & \multicolumn{2}{c}{\textbf{\textsc{te}}} & \multicolumn{2}{c}{\textbf{\textsc{Total}}} \\
            Metric &                   \o & $\sigma$ &                   \o & $\sigma$ &                   \o & $\sigma$ &                   \o & $\sigma$ &                   \o & $\sigma$ &                   \o & $\sigma$ &                   \o & $\sigma$ &                   \o & $\sigma$ &                      \o & $\sigma$ \\
\midrule
             \last &                 61.0 &      0.5 &                 43.3 &      1.3 &                 59.9 &      1.3 &                 69.7 &      0.6 &                 44.1 &      2.8 &                 59.0 &      0.9 &                 54.0 &      2.3 &                 42.6 &      7.3 &                    54.2 &      0.7 \\
           \srcdev &                 62.1 &      0.9 &                 44.2 &      1.7 &                 59.8 &      1.0 &                 69.2 &      0.8 &                 45.0 &      2.8 &                 59.3 &      0.8 &                 53.9 &      1.9 &                 44.6 &      4.7 &                    54.8 &      1.0 \\
           \trgdev &                 63.9 &      1.2 &                 45.7 &      2.8 &                 60.1 &      0.5 &                 71.5 &      0.3 &                 46.0 &      0.9 &                 60.2 &      0.9 &                 56.9 &      1.6 &                 47.8 &      3.7 &                    56.5 &      0.5 \\
\bottomrule
\end{tabular}
  \end{adjustbox}
  \caption{\zsxlt to TyDiQA-GoldP~\citep{clark-etal-2020-tydi}.}
\end{table*}

\begin{table*}[t]
 \footnotesize
  \begin{adjustbox}{width=\linewidth,center}
\begin{tabular}{l|cgcgcgcgcgcgcgcgcgcg|cg}
\toprule
\textbf{Languages} & \multicolumn{2}{c}{\textbf{\textsc{amh}}} & \multicolumn{2}{c}{\textbf{\textsc{hau}}} & \multicolumn{2}{c}{\textbf{\textsc{ibo}}} & \multicolumn{2}{c}{\textbf{\textsc{kin}}} & \multicolumn{2}{c}{\textbf{\textsc{lug}}} & \multicolumn{2}{c}{\textbf{\textsc{luo}}} & \multicolumn{2}{c}{\textbf{\textsc{pcm}}} & \multicolumn{2}{c}{\textbf{\textsc{swa}}} & \multicolumn{2}{c}{\textbf{\textsc{wol}}} & \multicolumn{2}{c}{\textbf{\textsc{yor}}} & \multicolumn{2}{c}{\textbf{\textsc{Total}}} \\
            Metric &                    \o & $\sigma$ &                    \o & $\sigma$ &                    \o & $\sigma$ &                    \o & $\sigma$ &                    \o & $\sigma$ &                    \o & $\sigma$ &                    \o & $\sigma$ &                    \o & $\sigma$ &                    \o & $\sigma$ &                    \o & $\sigma$ &                      \o & $\sigma$ \\
\midrule
             \last &                  30.7 &      1.2 &                  38.1 &      2.5 &                  13.6 &      4.1 &                  10.5 &      2.7 &                  12.0 &      3.5 &                   9.9 &      1.8 &                  39.6 &      2.1 &                  47.8 &      0.6 &                   9.7 &      2.2 &                  11.5 &      2.7 &                    22.3 &      1.8 \\
           \srcdev &                  30.3 &      0.8 &                  38.0 &      2.7 &                  14.4 &      5.1 &                  11.1 &      3.2 &                  13.0 &      4.4 &                  10.4 &      2.6 &                  40.1 &      2.2 &                  47.7 &      1.0 &                  10.3 &      3.1 &                  11.1 &      3.6 &                    22.6 &      2.4 \\
\bottomrule
\end{tabular}
  \end{adjustbox}
  \caption{\zsxlt to MasakhaNER \citep{adelani-etal-2021-masakhaner}.}
\end{table*}

\begin{table*}[t]
 \footnotesize
  \begin{adjustbox}{width=\linewidth,center}
\begin{tabular}{l|cgcgcgcgcgcgcgcgcgcgcgcgcg}
\toprule
\textbf{Languages} & \multicolumn{2}{c}{\textbf{\textsc{am}}} & \multicolumn{2}{c}{\textbf{\textsc{ar}}} & \multicolumn{2}{c}{\textbf{\textsc{ay}}} & \multicolumn{2}{c}{\textbf{\textsc{bg}}} & \multicolumn{2}{c}{\textbf{\textsc{de}}} & \multicolumn{2}{c}{\textbf{\textsc{el}}} & \multicolumn{2}{c}{\textbf{\textsc{es}}} & \multicolumn{2}{c}{\textbf{\textsc{fi}}} & \multicolumn{2}{c}{\textbf{\textsc{fr}}} & \multicolumn{2}{c}{\textbf{\textsc{he}}} & \multicolumn{2}{c}{\textbf{\textsc{hi}}} & \multicolumn{2}{c}{\textbf{\textsc{hu}}} & \multicolumn{2}{c}{\textbf{\textsc{ig}}} \\
            Metric &                   \o & $\sigma$ &                   \o & $\sigma$ &                   \o & $\sigma$ &                   \o & $\sigma$ &                   \o & $\sigma$ &                   \o & $\sigma$ &                   \o & $\sigma$ &                   \o & $\sigma$ &                   \o & $\sigma$ &                   \o & $\sigma$ &                   \o & $\sigma$ &                   \o & $\sigma$ &                   \o & $\sigma$ \\
\midrule
             \last &                 41.7 &      3.5 &                 42.8 &      2.2 &                 34.4 &      3.4 &                 78.2 &      0.4 &                 70.5 &      0.6 &                 73.5 &      0.4 &                 67.6 &      4.2 &                 73.5 &      0.6 &                 77.9 &      1.2 &                 53.4 &      0.5 &                 65.7 &      1.3 &                 74.4 &      0.8 &                 42.2 &      3.3 \\
\bottomrule
\end{tabular}
  \end{adjustbox}
  \caption{\zsxlt to WikiANN~\citep{pan-etal-2017-cross}.}
\end{table*}

\begin{table*}[t]
 \footnotesize
  \begin{adjustbox}{width=\linewidth,center}
\begin{tabular}{l|cgcgcgcgcgcgcgcgcgcgcg|cg}
\toprule
\textbf{Languages} & \multicolumn{2}{c}{\textbf{\textsc{ja}}} & \multicolumn{2}{c}{\textbf{\textsc{qu}}} & \multicolumn{2}{c}{\textbf{\textsc{ru}}} & \multicolumn{2}{c}{\textbf{\textsc{rw}}} & \multicolumn{2}{c}{\textbf{\textsc{sw}}} & \multicolumn{2}{c}{\textbf{\textsc{ta}}} & \multicolumn{2}{c}{\textbf{\textsc{te}}} & \multicolumn{2}{c}{\textbf{\textsc{tr}}} & \multicolumn{2}{c}{\textbf{\textsc{ur}}} & \multicolumn{2}{c}{\textbf{\textsc{vi}}} & \multicolumn{2}{c}{\textbf{\textsc{yo}}} & \multicolumn{2}{c}{\textbf{\textsc{Total}}} \\
            Metric &                   \o & $\sigma$ &                   \o & $\sigma$ &                   \o & $\sigma$ &                   \o & $\sigma$ &                   \o & $\sigma$ &                   \o & $\sigma$ &                   \o & $\sigma$ &                   \o & $\sigma$ &                   \o & $\sigma$ &                   \o & $\sigma$ &                   \o & $\sigma$ &                      \o & $\sigma$ \\
\midrule
             \last &                 16.7 &      1.0 &                 52.9 &      2.4 &                 65.8 &      1.6 &                 57.2 &      3.8 &                 61.6 &      1.7 &                 58.0 &      1.3 &                 51.0 &      1.5 &                 66.6 &      2.5 &                 53.1 &      7.4 &                 69.5 &      0.4 &                 29.8 &      4.1 &                    57.4 &      0.7 \\
           \srcdev &                 17.0 &      1.6 &                 53.9 &      1.7 &                 65.7 &      1.9 &                 56.7 &      3.8 &                 61.7 &      2.8 &                 57.9 &      1.6 &                 51.7 &      1.0 &                 66.4 &      2.7 &                 53.2 &      7.2 &                 69.7 &      1.3 &                 30.5 &      5.6 &                    57.7 &      0.9 \\
\bottomrule
\end{tabular}
  \end{adjustbox}
\end{table*}

\begin{table*}[t]
 \footnotesize
  \begin{adjustbox}{width=\linewidth,center}
\begin{tabular}{l|cgcgcgcgcgcgcgcgcgcgcg}
\toprule
\textbf{Languages} & \multicolumn{2}{c}{\textbf{\textsc{Afrikaans}}} & \multicolumn{2}{c}{\textbf{\textsc{Arabic}}} & \multicolumn{2}{c}{\textbf{\textsc{Basque}}} & \multicolumn{2}{c}{\textbf{\textsc{Bulgarian}}} & \multicolumn{2}{c}{\textbf{\textsc{Chinese}}} & \multicolumn{2}{c}{\textbf{\textsc{Dutch}}} & \multicolumn{2}{c}{\textbf{\textsc{Estonian}}} & \multicolumn{2}{c}{\textbf{\textsc{Finnish}}} & \multicolumn{2}{c}{\textbf{\textsc{French}}} & \multicolumn{2}{c}{\textbf{\textsc{German}}} & \multicolumn{2}{c}{\textbf{\textsc{Greek}}} \\
            Metric &                          \o & $\sigma$ &                       \o & $\sigma$ &                       \o & $\sigma$ &                          \o & $\sigma$ &                        \o & $\sigma$ &                      \o & $\sigma$ &                         \o & $\sigma$ &                        \o & $\sigma$ &                       \o & $\sigma$ &                       \o & $\sigma$ &                      \o & $\sigma$ \\
\midrule
             \last &                        86.8 &      0.4 &                     70.3 &      1.3 &                     55.4 &      1.9 &                        85.9 &      1.1 &                      29.3 &      6.4 &                    88.1 &      0.2 &                       80.5 &      1.3 &                      76.8 &      1.6 &                     75.5 &      1.1 &                     86.7 &      0.6 &                    57.8 &      1.6 \\
           \srcdev &                        86.8 &      0.5 &                     70.3 &      1.3 &                     55.4 &      1.9 &                        85.9 &      1.0 &                      29.5 &      6.4 &                    88.2 &      0.2 &                       80.5 &      1.3 &                      76.8 &      1.6 &                     75.4 &      1.0 &                     86.7 &      0.6 &                    57.7 &      1.6 \\
\bottomrule
\end{tabular}
  \end{adjustbox}
  \caption{\zsxlt to UDPOS as per XTREME benchmark $(1/2)$ \citep{hu2020xtreme}.}
\end{table*}

\begin{table*}[t]
 \footnotesize
  \begin{adjustbox}{width=\linewidth,center}
\begin{tabular}{l|cgcgcgcgcgcgcgcgcgcgcg}
\toprule
\textbf{Languages} & \multicolumn{2}{c}{\textbf{\textsc{Hebrew}}} & \multicolumn{2}{c}{\textbf{\textsc{Hindi}}} & \multicolumn{2}{c}{\textbf{\textsc{Hungarian}}} & \multicolumn{2}{c}{\textbf{\textsc{Indonesian}}} & \multicolumn{2}{c}{\textbf{\textsc{Italian}}} & \multicolumn{2}{c}{\textbf{\textsc{Japanese}}} & \multicolumn{2}{c}{\textbf{\textsc{Kazakh}}} & \multicolumn{2}{c}{\textbf{\textsc{Korean}}} & \multicolumn{2}{c}{\textbf{\textsc{Marathi}}} & \multicolumn{2}{c}{\textbf{\textsc{Persian}}} & \multicolumn{2}{c}{\textbf{\textsc{Portuguese}}} \\
            Metric &                       \o & $\sigma$ &                      \o & $\sigma$ &                          \o & $\sigma$ &                           \o & $\sigma$ &                        \o & $\sigma$ &                         \o & $\sigma$ &                       \o & $\sigma$ &                       \o & $\sigma$ &                        \o & $\sigma$ &                        \o & $\sigma$ &                           \o & $\sigma$ \\
\midrule
             \last &                     75.1 &      1.7 &                    67.3 &      1.7 &                        75.1 &      2.1 &                         71.5 &      0.2 &                      85.7 &      0.9 &                       21.6 &      4.5 &                     63.1 &      1.8 &                     36.8 &      1.8 &                      73.2 &      1.4 &                      66.9 &      1.2 &                         88.8 &      0.3 \\
           \srcdev &                     75.0 &      1.7 &                    67.4 &      1.8 &                        75.1 &      2.1 &                         71.5 &      0.2 &                      85.7 &      0.8 &                       21.7 &      4.5 &                     63.1 &      1.7 &                     36.9 &      1.8 &                      73.3 &      1.3 &                      66.8 &      1.2 &                         88.8 &      0.3 \\
\bottomrule
\end{tabular}
  \end{adjustbox}
\end{table*}

\begin{table*}[t]
 \footnotesize
  \begin{adjustbox}{width=\linewidth,center}
\begin{tabular}{l|cgcgcgcgcgcgcgcgcgcg|cg}
\toprule
\textbf{Languages} & \multicolumn{2}{c}{\textbf{\textsc{Russian}}} & \multicolumn{2}{c}{\textbf{\textsc{Spanish}}} & \multicolumn{2}{c}{\textbf{\textsc{Tagalog}}} & \multicolumn{2}{c}{\textbf{\textsc{Tamil}}} & \multicolumn{2}{c}{\textbf{\textsc{Telugu}}} & \multicolumn{2}{c}{\textbf{\textsc{Thai}}} & \multicolumn{2}{c}{\textbf{\textsc{Turkish}}} & \multicolumn{2}{c}{\textbf{\textsc{Urdu}}} & \multicolumn{2}{c}{\textbf{\textsc{Vietnamese}}} & \multicolumn{2}{c}{\textbf{\textsc{Yoruba}}} & \multicolumn{2}{c}{\textbf{\textsc{Total}}} \\
            Metric &                       \o & $\sigma$ &                        \o & $\sigma$ &                        \o & $\sigma$ &                      \o & $\sigma$ &                       \o & $\sigma$ &                     \o & $\sigma$ &                        \o & $\sigma$ &                     \o & $\sigma$ &                           \o & $\sigma$ &                       \o & $\sigma$ &                      \o & $\sigma$ \\
\midrule
             \last &                     83.0 &      0.7 &                      88.0 &      0.5 &                      88.8 &      1.2 &                    44.2 &      1.5 &                     70.6 &      1.9 &                   42.1 &      4.3 &                      59.9 &      1.8 &                   55.5 &      0.9 &                         57.5 &      0.4 &                     22.7 &      0.8 &                    66.6 &      0.8 \\
           \srcdev &                     83.0 &      0.7 &                      88.0 &      0.5 &                      88.9 &      0.9 &                    44.1 &      1.5 &                     70.6 &      1.8 &                   42.2 &      4.4 &                      59.9 &      1.8 &                   55.5 &      0.9 &                         57.5 &      0.3 &                     22.6 &      0.8 &                    66.6 &      0.9 \\
\bottomrule
\end{tabular}
  \end{adjustbox}
  \caption{\zsxlt to UDPOS as per XTREME benchmark $(1/2)$ \citep{hu2020xtreme}.}
\end{table*}

\begin{table*}[t]
\subsubsection{\fsxlt Results}
 \footnotesize
  \begin{adjustbox}{width=\linewidth,center}
\begin{tabular}{l|c|cgcgcgcgcgcgcg|cg}
\toprule
\textbf{Languages} & \textbf{Shots} & \multicolumn{2}{c}{\textbf{\textsc{aym}}} & \multicolumn{2}{c}{\textbf{\textsc{bzd}}} & \multicolumn{2}{c}{\textbf{\textsc{gn}}} & \multicolumn{2}{c}{\textbf{\textsc{hch}}} & \multicolumn{2}{c}{\textbf{\textsc{quy}}} & \multicolumn{2}{c}{\textbf{\textsc{shp}}} & \multicolumn{2}{c}{\textbf{\textsc{tar}}} & \multicolumn{2}{c}{\textbf{\textsc{Total}}} \\
            Metric &                &    \o & $\sigma$ &                    \o & $\sigma$ &                   \o & $\sigma$ &                    \o & $\sigma$ &                    \o & $\sigma$ &                    \o & $\sigma$ &                    \o & $\sigma$ &                      \o & $\sigma$ \\
\midrule
                   &                       5 &                  36.4 &      1.7 &                  38.2 &      2.0 &                 37.7 &      1.3 &                  36.2 &      2.0 &                  36.3 &      1.9 &                  38.6 &      2.4 &                  35.4 &      1.5 &                    37.0 &      0.3 \\
                   &                      10 &                  37.8 &      2.4 &                  39.9 &      2.8 &                 40.2 &      3.2 &                  37.5 &      2.5 &                  37.5 &      2.4 &                  39.4 &      2.2 &                  37.6 &      3.4 &                    38.6 &      0.4 \\
             \last &                      50 &                  42.9 &      1.9 &                  45.3 &      2.2 &                 45.1 &      2.3 &                  41.9 &      1.6 &                  44.1 &      3.0 &                  45.4 &      1.8 &                  42.4 &      2.5 &                    43.9 &      0.3 \\
\bottomrule
\end{tabular}
  \end{adjustbox}
  \caption{Multilingual \fsxlt to 7 languages of AmNLI~\citep{ebrahimi-americasnli}.}
\end{table*}

\begin{table*}[t]
 \footnotesize
  \begin{adjustbox}{width=\linewidth,center}
\begin{tabular}{l|c|cgcgcgcgcgcgcgcg|cg}
\toprule
\textbf{Languages} & \textbf{Shots} & \multicolumn{2}{c}{\textbf{\textsc{ar}}} & \multicolumn{2}{c}{\textbf{\textsc{bn}}} & \multicolumn{2}{c}{\textbf{\textsc{fi}}} & \multicolumn{2}{c}{\textbf{\textsc{id}}} & \multicolumn{2}{c}{\textbf{\textsc{ko}}} & \multicolumn{2}{c}{\textbf{\textsc{ru}}} & \multicolumn{2}{c}{\textbf{\textsc{sw}}} & \multicolumn{2}{c}{\textbf{\textsc{te}}} & \multicolumn{2}{c}{\textbf{\textsc{Total}}} \\
            Metric &               &    \o & $\sigma$ &                   \o & $\sigma$ &                   \o & $\sigma$ &                   \o & $\sigma$ &                   \o & $\sigma$ &                   \o & $\sigma$ &                   \o & $\sigma$ &                   \o & $\sigma$ &                      \o & $\sigma$ \\
\midrule
                   &                       5 &                 61.3 &     12.4 &                 49.1 &      2.2 &                 61.6 &      1.3 &                 72.2 &      1.1 &                 51.8 &      2.1 &                 61.1 &      2.1 &                 60.0 &      1.7 &                 46.4 &      4.2 &                    57.9 &      0.8 \\
                   &                      10 &                 65.3 &      2.8 &                 51.1 &      2.6 &                 62.8 &      1.7 &                 72.7 &      1.1 &                 53.0 &      3.0 &                 62.3 &      1.4 &                 61.5 &      1.7 &                 54.9 &      5.0 &                    60.4 &      0.3 \\
             \last &                      50 &                 69.7 &      1.1 &                 59.3 &      3.3 &                 66.7 &      2.0 &                 74.6 &      1.0 &                 57.4 &      2.0 &                 64.3 &      0.8 &                 68.3 &      1.4 &                 68.1 &      3.3 &                    66.0 &      0.2 \\
\bottomrule
\end{tabular}
  \end{adjustbox}
  \caption{Multilingual \fsxlt to 8 languages of TyDiQA-GoldP~\citep{clark-etal-2020-tydi}.}
\end{table*}

\begin{table*}[t]
 \footnotesize
  \begin{adjustbox}{width=\linewidth,center}
\begin{tabular}{l|c|cgcgcgcgcgcgcgcgcg|cg}
\toprule
\textbf{Languages} & \textbf{\textsc{shots}} & \multicolumn{2}{c}{\textbf{\textsc{ar}}} & \multicolumn{2}{c}{\textbf{\textsc{fi}}} & \multicolumn{2}{c}{\textbf{\textsc{hu}}} & \multicolumn{2}{c}{\textbf{\textsc{sw}}} & \multicolumn{2}{c}{\textbf{\textsc{ta}}} & \multicolumn{2}{c}{\textbf{\textsc{tr}}} & \multicolumn{2}{c}{\textbf{\textsc{ur}}} & \multicolumn{2}{c}{\textbf{\textsc{vi}}} & \multicolumn{2}{c}{\textbf{\textsc{zh}}} & \multicolumn{2}{c}{\textbf{\textsc{Total}}} \\
            Metric &    {} &               \o & $\sigma$ &                   \o & $\sigma$ &                   \o & $\sigma$ &                   \o & $\sigma$ &                   \o & $\sigma$ &                   \o & $\sigma$ &                   \o & $\sigma$ &                   \o & $\sigma$ &                   \o & $\sigma$ &                      \o & $\sigma$ \\
\midrule
                   &                       5 &                 68.6 &      2.2 &                 76.2 &      1.2 &                 77.4 &      1.0 &                 69.2 &      6.5 &                 63.5 &      2.8 &                 73.1 &      3.2 &                 71.0 &      3.2 &                 74.9 &      1.8 &                 34.1 &      5.2 &                    67.6 &      0.1 \\
                   &                      10 &                 70.0 &      2.1 &                 77.4 &      1.0 &                 78.8 &      0.7 &                 77.2 &      5.3 &                 66.9 &      2.0 &                 77.1 &      2.0 &                 76.9 &      2.2 &                 75.9 &      1.5 &                 37.0 &      4.7 &                    70.8 &      0.4 \\
\bottomrule
\end{tabular}
  \end{adjustbox}
  \caption{Multilingual \fsxlt to 9 languages WikiANN~\citep{pan-etal-2017-cross}.}
\end{table*}

\begin{table*}[t]
 \footnotesize
  \begin{adjustbox}{width=\linewidth,center}
\begin{tabular}{l|c|cgcgcgcgcgcgcgcgcg|cg}
\toprule
\textbf{Languages} & \textbf{\textsc{shots}} & \multicolumn{2}{c}{\textbf{\textsc{Arabic}}} & \multicolumn{2}{c}{\textbf{\textsc{Basque}}} & \multicolumn{2}{c}{\textbf{\textsc{Chinese}}} & \multicolumn{2}{c}{\textbf{\textsc{Finnish}}} & \multicolumn{2}{c}{\textbf{\textsc{German}}} & \multicolumn{2}{c}{\textbf{\textsc{Indonesian}}} & \multicolumn{2}{c}{\textbf{\textsc{Japanese}}} & \multicolumn{2}{c}{\textbf{\textsc{Turkish}}} & \multicolumn{2}{c}{\textbf{\textsc{Urdu}}} & \multicolumn{2}{c}{\textbf{\textsc{Total}}} \\
            Metric &              {}  &     \o & $\sigma$ &                       \o & $\sigma$ &                        \o & $\sigma$ &                        \o & $\sigma$ &                       \o & $\sigma$ &                           \o & $\sigma$ &                         \o & $\sigma$ &                        \o & $\sigma$ &                     \o & $\sigma$ &                      \o & $\sigma$ \\
\midrule
                   &                       5 &                     81.3 &      1.2 &                     72.8 &      1.5 &                      65.8 &      1.6 &                      83.8 &      0.6 &                     88.8 &      0.3 &                         73.1 &      0.6 &                       76.0 &      2.3 &                      69.3 &      1.6 &                   80.7 &      1.1 &                    76.8 &      0.1 \\
                   &                      10 &                     83.4 &      0.6 &                     76.4 &      1.1 &                      68.7 &      1.8 &                      84.8 &      0.4 &                     89.6 &      0.4 &                         74.3 &      0.3 &                       79.0 &      1.0 &                      72.7 &      0.9 &                   83.6 &      0.7 &                    79.2 &      0.1 \\
\bottomrule
\end{tabular}
  \end{adjustbox}
  \caption{Multilingual \fsxlt to 9 languages of UDPOS~\citep{zeman-2020-ud,hu2020xtreme}.}
\end{table*}

\end{document}